\documentclass{article} 
\usepackage{iclr2019_conference,times}

\usepackage[utf8]{inputenc} 
\usepackage[T1]{fontenc}    
\usepackage{url}            
\usepackage{booktabs}       
\usepackage{amsfonts}       
\usepackage{nicefrac}       
\usepackage{microtype}      
\usepackage{todonotes}      

\usepackage{amssymb}
\usepackage[ruled,lined]{algorithm2e}
\usepackage{graphicx}
\usepackage[lofdepth,lotdepth]{subfig}
\usepackage{amsthm}
\usepackage{color}
\usepackage{amsmath}
\usepackage{bigints}
\usepackage{multirow}
\usepackage{arydshln}
\usepackage{todonotes}

\theoremstyle{plain}
\newcounter{theoremcounter}
\newtheorem{theorem}[theoremcounter]{Theorem}

\theoremstyle{definition}

\usepackage{color}

\title{MAE: Mutual Posterior-Divergence Regularization for Variational AutoEncoders}

\author{Xuezhe Ma, Chunting Zhou \& Eduard Hovy \\
	Carnegie Mellon University \\
	\texttt{\{xuezhem, ctzhou, ehovy\}@cs.cmu.edu}
}

\iclrfinalcopy 

\begin{document}

\maketitle

\begin{abstract}
	Variational Autoencoder (VAE), a simple and effective deep generative model, has led to a number of impressive empirical successes and spawned many advanced variants and theoretical investigations. However, recent studies demonstrate that, when equipped with expressive generative distributions (aka. decoders), VAE suffers from learning uninformative latent representations with the observation called \emph{KL Varnishing}, in which case VAE collapses into an unconditional generative model. In this work, we introduce \emph{mutual posterior-divergence regularization}, a novel regularization that is able to control the geometry of the latent space to accomplish meaningful representation learning, while achieving comparable or superior capability of density estimation.
	Experiments on three image benchmark datasets demonstrate that, when equipped with powerful decoders, our model performs well both on density estimation and representation learning.
\end{abstract}

\section{Introduction}

Representation learning, besides data distributions estimation, is a principle component in generative models. The goal is to identify and disentangle the underlying causal factors, to tease apart the underlying dependencies of the data, so that it becomes easier to understand, to classify, or to perform other tasks~\citep{bengio2013representation}.
Among these generative models,  VAE~\citep{kingma2014auto,rezende14} gains popularity for its capability of estimating densities of complex distributions, while automatically 
learning meaningful (low-dimensional) representations from raw data.
VAE, as a member of latent variable models (LVMs), defines the joint distribution between the observed data (visible variables) and a set of latent variables by factorizing it as the product of a prior over the latent variables and a conditional distribution of the visible variables given the latent ones (detailed in \S\ref{sec:vae}).
VAEs are usually estimated by maximizing the likelihood of the observed data by marginalizing over the latent variables, typically via optimizing the evidence lower bound (ELBO).
By learning a VAE from the data with the appropriate hierarchical structure of latent variables, the hope is to uncover and untangle the causal sources of variations that we are interested in. 

A notorious problem of VAEs, however, is that the marginal likelihood may not guide the model to learn the intended latent variables. It may instead focus on explaining irrelevant but common correlations in the data~\citep{ganchev2010posterior}.
Extensive previous studies~\citep{bowman2015generating,chenvariational,yang2017improved} showed that optimizing the ELBO objective is often completely disconnected from the goal of learning good representations.
An extreme case called \mbox{\emph{KL varnishing}}, happens when using sufficiently expressive decoding distributions such as auto-regressive ones; the latent variables are often completely ignored and the VAE regresses to a standard auto-regressive model~\citep{larochelle2011neural,oord2016pixel}.

This problem has spawned significant interests in analyzing and solving it from both theoretical and practical perspectives. We can only name a few here due to space limits.
Some previous work~\citep{bowman2015generating,sonderby2016train,serban2017hierarchical} attributed the KL varnishing phenomenon to ``optimization challenges'' of VAEs, and proposed training methods including annealing the relative weight of the KL term in ELBO~\citep{bowman2015generating,sonderby2016train} or adding \emph{free bits}~\citep{kingma2016improved,chenvariational}. 
However, \citet{chenvariational} pointed out that this phenomenon arises not just due to the optimization challenges, and even if we find the exact solution for the optimization problems, the latent code will still be ignored at optimum. 
They proposed a solution by limiting the capacity of the decoder and applied PixelCNN~\citep{oord2016pixel} with small local receptive fields as the decoder of VAEs to model 2D images, achieving both impressive performance for density estimation and informative latent representations. 
\citet{yang2017improved} embraced a similar idea and applied VAE to text modeling by using dilated CNN as the decoder.
Unfortunately, these approaches require manual and problem-specific design of the decoder's architecture to learn meaningful representations. 
Other studies attempted to explore alternatives of ELBO.
\citet{makhzani2015adversarial} proposed Adversarial Autoencoders (AAEs) by replacing the KL-divergence between the posterior and prior distributions with Jensen-Shannon divergence on the aggregated posterior distribution. 
InfoVAEs~\citep{zhao2017infovae} generalized the Jensen-Shannon divergence in AAEs to a divergence family and linked its objective to the Mutual Information between the data and the latent variables.
However, directly optimizing these objectives is intractable, requiring advanced approximate learning methods such as adversarial learning or Maximum-Mean Discrepancy~\citep{gretton2007kernel,dziugaite2015training,li2015generative}.
Moreover, these models' performance on density estimation significantly falls behind state-of-the-art models~\citep{salimans2017pixelcnn++,chenvariational}.

In this paper, we propose to tackle the aforementioned representation learning challenges of VAEs by adding a data-dependent regularization to the ELBO objective.
Our contributions are three-fold: (1) Algorithmically, we introduce the \emph{mutual posterior-divergence regularization} for VAEs, named MAEs (\S\ref{subsec:mae}),
to control the geometry of the latent space during learning by encouraging the learned variational posteriors to be diverse (i.e. they are favored to be mutually ``different'' from each other), to achieve low-redundant, interpretable representation learning.
(2) Theoretically, we establish a close relation between MAE and InfoVAE, by showing that the mutual posterior-devergence regularization maximizes a symmetric version of the KL divergence involved in InfoVAE's mutual information term (\S\ref{subsec:infovae}). 
(3) Experimentally, on three benchmark datasets for images, we demonstrate the effectiveness of MAE as a density estimator by state-of-the-art log-likelihood results on MNIST and OMNIGLOT, and comparable result on CIFAR-10.
Moreover, by performing image reconstruction, unsupervised and semi-supervised classification, we show that MAE is also capable of learning meaningful latent representations, even combined with a sufficiently powerful decoder (\S\ref{sub:experiment}).

\section{Variational Autoencoders}\label{sec:vae}

\subsection{Notations}\label{subsec:notation}
Throughout we use uppercase letters for random variables, and lowercase letters for realizations of the corresponding random variables. 
Let $X \in \mathcal{X}$ be the randoms variables of the observed data, e.g., $X$ is an image or a sentence for image and text generation, respectively.

Let $P$ denote the true distribution of the data, i.e., $X \sim P$, and $D = \{x_1, \ldots, x_N\}$ be our training sample, where $x_i, i=1,\ldots, N,$
are usually i.i.d. samples of $X$.
Let $\mathcal{P} = \{P_\theta : \theta \in \Theta\}$ denote a parametric statistical model indexed by parameter $\theta \in \Theta$, where $\Theta$ is the parameter space.
$p$ is used to denote the density of corresponding distribution $P$.
In the literature of deep generative models, deep neural networks are the most widely used parametric models.
The goal of generative models is to learn the parameter $\theta$ such that $P_{\theta}$ can best approximate the true distribution $P$.

\subsection{VAEs}\label{subsec:vae}
In the framework of VAEs, or general LVMs, a set of latent variables $Z \in \mathcal{Z}$ are introduced to characterize the hidden patterns of $X$, 
and the model distribution $P_{\theta}(X)$ is defined as the marginal of the joint distribution between $X$ and $Z$:
\begin{equation}\label{eq:distr}
p_{\theta}(x) = \int_{\mathcal{Z}} p_{\theta}(x, z) d\mu(z) = \int_{\mathcal{Z}} p_{\theta}(x|z) p_{\theta}(z) d\mu(z), \quad \forall x \in \mathcal{X},
\end{equation}
where the joint distribution $p_{\theta}(x, z)$ is factorized as the product of a prior $p_{\theta}(z)$ over the latent $Z$, and the ``generative'' distribution $p_{\theta}(x|z)$. $\mu(z)$ is the base measure on the latent space $\mathcal{Z}$.
Typically, prior $p_{\theta}(z)$ is modeled with a simple distribution like multivariate Gaussian, or transforming simple priors to complex ones by normalizing flows and variants~\citep{rezende2015variational,kingma2016improved,sonderby2016ladder}. 

To learn parameters $\theta$, we wish to minimizes the negative log-likelihood of the parameters:
\begin{equation}\label{eq:mle}
\min\limits_{\theta \in \Theta} \frac{1}{N} \sum\limits_{i=1}^{N} -\log p_{\theta}(x_i) = \min\limits_{\theta \in \Theta} \mathrm{E}_{\tilde{P}(X)} [-\log p_{\theta}(X)]
\end{equation}
where $\tilde{P}(X)$ is the empirical distribution derived from training data $D$.
In general, this marginal likelihood is intractable to compute or differentiate directly for high-dimensional latent space $\mathcal{Z}$.
Variational Inference~\citep{wainwright2008graphical} provides a solution to optimize the \emph{evidence lower bound} (ELBO) an alternative objective by introducing a parametric \emph{inference model} $q_{\phi}(z|x)$:
\begin{equation}\label{eq:elbo}
{\arraycolsep=2pt\def\arraystretch{1.5}
\begin{array}{rcl}
\mathcal{L}_{elbo}(\theta, \phi) & = & \mathrm{E}_{p(X)}\left[ -\mathrm{E}_{q_{\phi} (Z|X)} [\log p_{\theta} (X|Z)] + \mathrm{KL}(q_{\phi} (Z|X) || p_{\theta}(Z)) \right] \\
 & = & \mathrm{E}_{p(X)}\left[-\log p_{\theta}(X) + \mathrm{KL}(q_{\phi} (Z|X) || p_{\theta}(Z|X)) \right] \geq \mathrm{E}_{p(X)}\left[-\log p_{\theta}(X) \right]
\end{array}
}
\end{equation}
where $\mathcal{L}_{elbo}$ could be seen as an autoencoding loss with $q_{\phi} (z|x)$ being the encoder and $p_{\theta} (x|z)$ being the decoder, with the first term in the RHS in \eqref{eq:elbo} as the reconstruction error.

\subsection{Autoencoding Problem in VAEs}
As discussed in \citet{chenvariational}, without further assumptions, the ELBO objective $\mathcal{L}_{elbo}$ in \eqref{eq:elbo} may not guide the model towards the intended role for the latent variables $Z$, or even learn uninformative $Z$ with the observation that the KL term $\mathrm{KL}(q_{\phi} (Z|X) || p_{\theta}(Z))$ varnishes to zero.
For example, suppose we use an auto-regressive decoder, $p_{\theta}(x|z) = \prod_{i} p_{\theta}(x_i| x_{<i}, z)$, which is sufficiently expressive that it can model the data distribution $P(X)$ without the assistance of $Z$, i.e, $p_{\theta}(x_i| x_{<i}, z) = p_{\theta}(x_i| x_{<i})$. 
In this case, the optimal $Z$ w.r.t. $\mathcal{L}_{elbo}$ is the one independent with $X$, with the inference model reducing to the prior, i.e., $q_{\phi} (Z|X) = p_{\theta}(Z), \forall X \in \mathcal{X}$.

The essential reason of this problem is that, under absolutely unsupervised setting, the marginal likelihood based objective $\mathcal{L}_{elbo}$ incorporates no (direct) supervision on the latent space to characterize the latent variable $Z$ with preferred properties w.r.t. representation learning. The main goal of this work is to explicitly control the geometry of the latent space, in the hope that preferred latent representations would be characterized and selected. 

\section{Mutual Posterior-Divergence Regularization}\label{sec:mae}
\subsection{Geometric Properties of Meaningful Latent Space}
Motivated by the Diversity-Inducing Mutual Angular Regularization~\citep{xie2015latent} which is widely used in LVMs, we propose to regularize the posteriors $q_{\phi}(z|x)$ of different data $x \in \mathcal{X}$, to encourage them to diversely, smoothly, and evenly spread out in the data space $\mathcal{X}$. The intuition is:
(1) To make posteriors mutually diverse from each other, the patterns captured by different posteriors are likely to have less redundancy and hence characterizing and interpreting different data $x$.
(2) To make posteriors smoothly and evenly distributed in the whole space of $\mathcal{X}$, the shared patterns of similar data points are likely to be captured by their posteriors, to avoid isolating each data point from others.
By balancing the diversity and smoothness of the distribution of posteriors, learned representations are encouraged to maintain global structured information, discarding detailed texture of local dependencies in the data.

\subsection{MAEs}\label{subsec:mae}
\paragraph{Measure of Diversity.} We propose to use expectation of the mutual KL-divergence between a pair of data to measure the diversity of posteriors. Specifically, the mutual posterior diversity is defined as:
\begin{equation}\label{eq:mpd}
MPD = \mathrm{E}_{X_1, X_2\sim P(X)} [\mathrm{KL}(q_{\phi}(Z|X_1) || q_{\phi}(Z|X_2))]
\end{equation}
There are two main reasons we use KL-divergence instead of others as the measure of diversity: (1) KL-divergence is transformation invariant, i.e., for an invertible smooth function $f$,
\begin{displaymath}
\mathrm{KL}(q_{\phi}(Z|X_1) || q_{\phi}(Z|X_2)) = \mathrm{KL}(q_{\phi}(f(Z)|X_1) || q_{\phi}(f(Z)|X_2))
\end{displaymath}
It makes the computation efficient for complex posteriors that are transformed from simple ones, such as applying normalizing flows and variants~\citep{rezende2015variational,kingma2016improved,sonderby2016ladder}. 
(2) KL-divergence has a close relation with mutual information, an important information-theoretic measure of the mutual dependence between two variables, which provides us the theoretical justification of the proposed regularizer (detailed in \S\ref{subsec:infovae}).

\begin{figure}[t]
\centering
\includegraphics[scale=0.28]{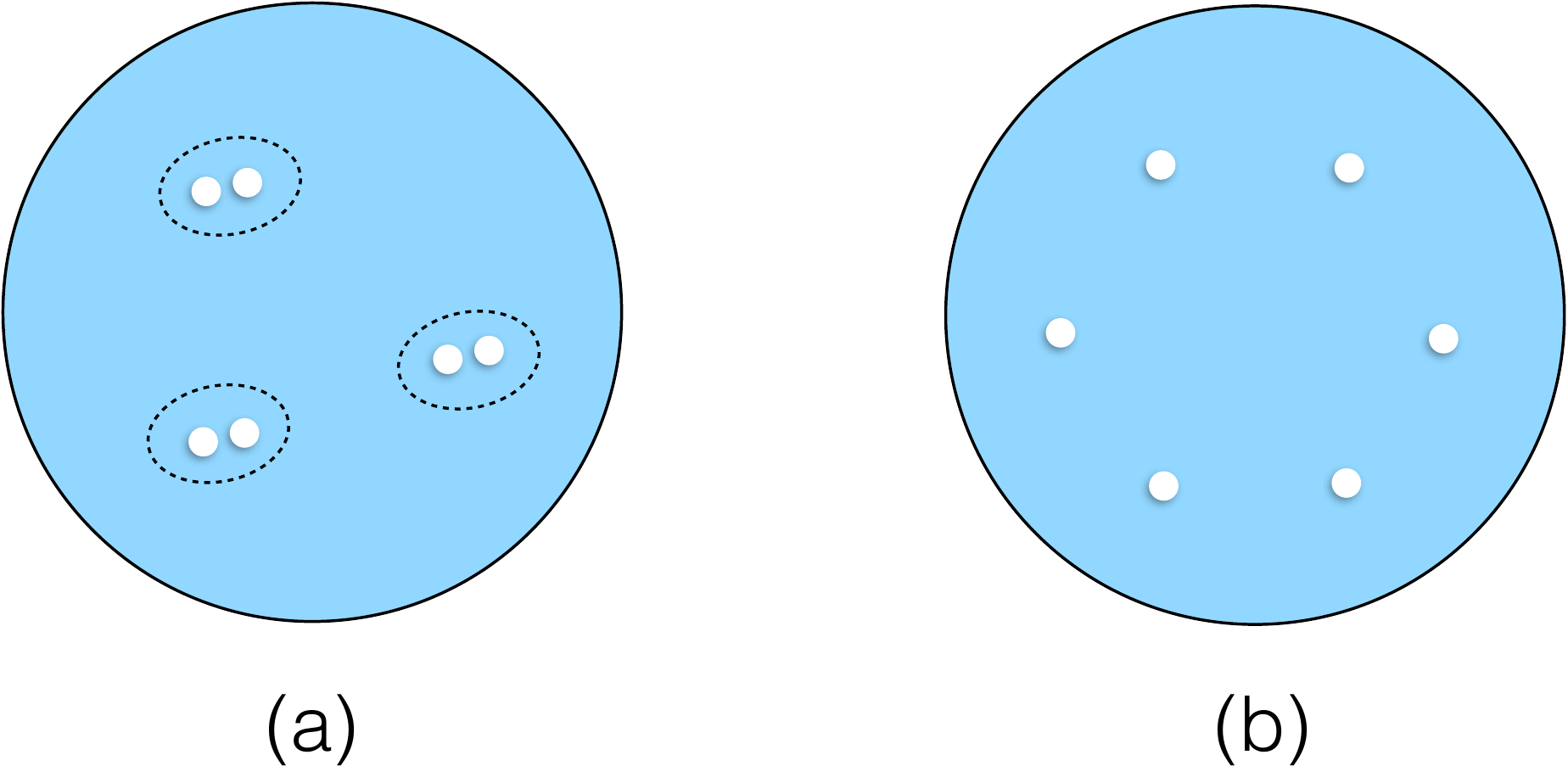}
\caption{The mean of pairwise distances between points in (a) is close to (b), while the standard deviation in (a) is much larger.}
\label{fig:std}
\vspace{-5pt}
\end{figure}

In MAEs, we propose to maximize the mutual posterior diversity (MPD) in \eqref{eq:mpd}. A straightforward way is to add negative MPD to the objective $\mathcal{L}_{elbo}$ in \eqref{eq:elbo} that VAEs attempt to minimize. 
There are, however, two practical issues: (1) The scale of MPD, particularly for continuous $Z$, is much larger than that of $\mathcal{L}_{elbo}$. We need to choose a hyperparameter carefully to control the scale of MPD, making optimization much more challenging and unstable.
(2) For multivariate $Z$, e.g. a $K$-dimensional $Z = (Z_1, Z_2, \ldots, Z_K)$, due to the property of KL-divergence, MPD may be dominated by a small group of dimensions, leaving others close to zero. In this case, most dimensions of $Z$ are uninformative, which is not a desired representation.

To solve the two problems, in practice, we propose to minimize a MPD-based loss instead of directly maximizing MPD itself:
\begin{equation}
\mathcal{L}_{diverse} = \mathrm{E}_{X_1, X_2\sim P(X)} \left[\sum\limits_{k=1}^{K} \log(1 + \exp(-\mathrm{KL}(q_{\phi}(Z_k|X_1) || q_{\phi}(Z_k|X_2)))) \right]
\end{equation}
$\mathcal{L}_{diverse}$ has two important good properties: (1) $\mathcal{L}_{diverse} \geq 0$. (2) $\mathcal{L}_{diverse} \to 0$ iff $\mathrm{KL}(q_{\phi}(Z_k|X_1) || q_{\phi}(Z_k|X_2)) \to \infty, \forall k$.
The first property sets a lower bound of $\mathcal{L}_{diverse}$, making optimization much more stable. The second one guarantees that all the dimensions of the latent $S$ need to be mutually diverse w.r.t minimizing $\mathcal{L}_{diverse}$. 

\paragraph{Measure of Smoothness.} The smoothness of the distribution of posteriors is measured by utilizing the standard deviation of the mutual KL-divergence:
\begin{equation}\label{eq:std}
\mathcal{L}_{smooth} = \mathrm{STD}_{X_1, X_2\sim P(X)} [\mathrm{KL}(q_{\phi}(Z|X_1) || q_{\phi}(Z|X_2))]
\end{equation}
where $\mathrm{STD}$ stands for standard deviation of random variables.

$\mathcal{L}_{smooth}$ encourages the posteriors to smoothly and evenly spread out to different directions.
Encouraging the standard deviation to be small can prevent the phenomenon that the posteriors fall into several small groups that are isolated from each other. 
It is crucially important for unsupervised clustering tasks, in which we want to cluster similar data into a big group instead of splitting them into multiple separated small groups (see \S\ref{subsec:binary} for detailed experimental results).
Figure~\ref{fig:std} shows two sets of distributions of data points, where the mean of the pairwise distances of the first set (Figure~\ref{fig:std} (a)) is roughly the same as the second set (Figure~\ref{fig:std} (b)). 
But the standard deviation of the first set is larger.

In the framework of MAEs, the final objective to minimize is:
\begin{equation}\label{eq:mae}
\mathcal{L}_{MAE} = \mathcal{L}_{elbo} + \eta \, \mathcal{L}_{diverse} + \gamma \, \mathcal{L}_{smooth}
\end{equation}
where $\eta > 0, \gamma > 0$ are regularization constants to balance the three losses in $\mathcal{L}_{MAE}$. 
Even though MAE introduces two extra hyperparameters $\eta$ and $\gamma$, we find them easy to tune and MAE shows robust performance with different values of $\eta$ and $\gamma$.

To solve \eqref{eq:mae}, we can approximate $\mathcal{L}_{diverse}$ and $\mathcal{L}_{smooth}$ using Monte carlo in each mini-batch:
\begin{equation}\label{eq:approx}
\mathcal{L}_{diverse} \approx \frac{1}{M} \sum\limits_{x_1 \neq x_2} \sum\limits_{k=1}^{K} \log(1 + \exp(-\mathrm{KL}(q_{\phi}(Z_k|x_1) || q_{\phi}(Z_k|x_2))))
\end{equation}
where $M$ is the number of valid pairs of data in each mini-batch. $\mathcal{L}_{smooth}$ is approximately computed similarly.

\subsection{Theoretical Justification}\label{subsec:infovae}
So far our discussion has been concentrated on the motivation and mathematical formulation of the proposed regularization method for VAE. In this section, we provide theoretical justification by connecting the mutual posterior diversity (MPD) in \eqref{eq:mpd} with the mutual information term defined in InfoVAE~\citep{zhao2017infovae}. 
With the end goal of theoretically justifying the proposed regularizer in mind, we first review the background of the mutual information (MI) term involved in the InfoVAE objective, which is central for linking MAE and InfoVAE.

\paragraph{Mutual Information Maximization.} InfoVAE proposed the mutual information by first defining the joint ``inference distribution'':
\begin{displaymath}
q_{\phi}(x, z) = p(x) q_{\phi}(z|x)
\end{displaymath}
where $p(x)$ is the density of the true data distribution $P(X)$. Then they added a mutual information maximization term that prefers high mutual information between $X$ and $Z$ under $q_{\phi}(x, z)$ to the standard $\mathcal{L}_{elbo}$:
\begin{displaymath}
\mathcal{L}_{InfoVAE} = \mathcal{L}_{elbo} - \lambda I_{q_{\phi}(x, z)} (x;z)
\end{displaymath}
and further proved that 
\begin{equation}\label{eq:mi}
I_{q_{\phi}(x, z)} (x;z) = \mathrm{E}_{P(X)} [\mathrm{KL} (q_{\phi}(z|x) || q_{\phi}(z))]
\end{equation}
where $q_{\phi}(z) = \int_{\mathcal{X}} q_{\phi}(x, z) d\mu(x)$ is the marginal of $q_{\phi}(x, z)$. 
Mutual information inspired objectives have been explored in GANs~\citep{goodfellow2014generative,chen2016infogan}, clustering~\citep{hinton1995wake,krause2010discriminative} and representation learning~\citep{esmaeili2018structured,hjelm2018learning}.

\paragraph{Relation between MPD and MI.} The following theorem states our major result that reveals the relation between MPD and MI (proof in Appendix~\ref{appendix:proof:thm1}):
\begin{theorem}\label{thm:mae}
The mutual posterior diversity (MPD) in \eqref{eq:mpd} is a symmetric version of the KL-divergence of MI in \eqref{eq:mi}:
\begin{equation}
MPD = \mathrm{E}_{P(X)} [\mathrm{KL} (q_{\phi}(z|x) || q_{\phi}(z)) + \mathrm{KL} (q_{\phi}(z) || q_{\phi}(z|x))]
\end{equation}
\end{theorem}
Roughly, Theorem~\ref{thm:mae} states that maximizing MPD and MI achieve the same goal: maximizing the divergence between the posterior distribution $q_{\phi}(z|x)$ and the marginal $q_{\phi}(z)$. 
Note that the (approximate) computation of MPD, as described in \eqref{eq:approx}, is much easier than MI, which is generally intractable and requires adversarial learning or Maximum-Mean Discrepancy.

\section{Experiments}\label{sub:experiment}
In this paper, we choose Variational Lossy Autoencoder (VLAE)~\citep{chenvariational}, VAE with auto-regressive flow (AF) prior, and auto-regressive decoder, as the basic architecture of our MAE models. 
More detailed descriptions, results, and analysis of the conducted experiments are provided in Appendix~\ref{appendix:experiment}.

\subsection{Binary Images}\label{subsec:binary}
We evaluate MAE on two binary images that are commonly used for evaluating deep generative models: MNIST~\citep{lecun1998gradient} and OMNIGLOT~\citep{lake2013one,burda2015importance}, both with dynamically binarized version~\citep{burda2015importance}. 
VLAE networks used in binary image datasets are similar of that described in \citet{chenvariational}: ResNet~\citep{he2016deep} encoder same as in ResNet VAE~\citep{kingma2016improved}, PixelCNN~\citep{oord2016pixel} decoder with 6 layers of masked convolution, and 32-dimensional latent code with AF prior implemented with MADE~\citep{germain2015made}. 
The only difference is that the PixelCNN decoder has varying filter sizes: two 7x7 layers, followed by two 5x5 layers, and finally two 3x3 layers, instead of a fixed filter size of 3x3 used in ~\citet{chenvariational}. 
Hence the decoder we use has larger receptive field, to ensure that the decoder is sufficiently expressive.
The same architecture is applied to all the experiments on both the two datasets.
For pair comparison, we re-implemented VLAE using the same architecture in our MAE model.
``Free bits''~\citep{kingma2016improved} is used to improve optimization stability of VLAE (not for MAE).
For hyperparameters $\eta$ and $\gamma$, we explored a few configurations: $\eta$ is selected from $[0.5, 1.0, 2.0]$, and $\gamma$ from $[0.1, 0.5, 1.0]$.

\begin{table}
\caption{Image modeling results on dynamically binarized MNIST and OMNIGLOT.}
\label{tab:density}
{\small
\begin{minipage}[t]{0.49\textwidth}
\centering
\subfloat[MNIST]{
\begin{tabular}[t]{l|c}
\hline
\textbf{Model} & \textbf{NLL}(\textbf{KL}) \\
\hline
IWAE~\citep{burda2015importance} & 82.90 \\
LVAE~\citep{sonderby2016ladder} & 81.74 \\
InfoVAE~\citep{zhao2017infovae} & 80.76 \\
Discrete VAE~\citep{rolfe2016discrete} & 80.04 \\
IAF VAE~\citep{kingma2016improved} & 79.10 \\
VLAE~\citep{chenvariational} & 78.53 \\
VLAE (re-impl) & 78.26 (9.02) \\
\hline
MAE: $\eta=1.0, \gamma=0.1$ & 78.02 (11.38) \\
MAE: $\eta=0.5, \gamma=0.5$ & 78.00 (10.44) \\
MAE: $\eta=1.0, \gamma=0.5$ & \textbf{77.98 (11.54)} \\
MAE: $\eta=2.0, \gamma=0.5$ & 77.99 (12.67) \\
MAE: $\eta=1.0, \gamma=1.0$ & 78.15 (10.19) \\
\hline
\end{tabular}
}
\end{minipage}
\begin{minipage}[t]{0.49\textwidth}
\centering
\subfloat[OMNIGLOT]{
\begin{tabular}[t]{l|c}
\hline
\textbf{Model} & \textbf{NLL}(\textbf{KL}) \\
\hline
IWAE~\citep{burda2015importance} & 103.38 \\
LVAE~\citep{sonderby2016ladder} & 102.11 \\
Discrete VAE~\citep{rolfe2016discrete} & 97.43 \\
SA-VAE~\citep{kim2018semi} & 90.05 (2.78) \\
VLAE~\citep{chenvariational} & 89.83 \\
VampPrior~(Tomczak, \citeyear{tomczak2018vae}) & 89.76 \\
VLAE (re-impl) & 89.62 (8.43) \\
\hline
MAE: $\eta=0.5, \gamma=0.1$ & 89.21 (9.08) \\
MAE: $\eta=0.5, \gamma=0.2$ & \textbf{89.09 (12.66)} \\
MAE: $\eta=1.0, \gamma=0.2$ & 89.15 (14.86) \\
MAE: $\eta=0.5, \gamma=0.5$ & 89.41 (10.53) \\
\hline
\end{tabular}
}
\end{minipage}
}
\end{table}

\begin{table}[t]
\caption{Performance of unsupervised clustering and semi-supervised classification. For each experiment, we report the average over 5 runs.}
\label{tab:classify}
\centering
{\small
\begin{tabular}[t]{l|ccc|ccc:ccc}
\hline
 & \multicolumn{3}{c|}{unsupervised clustering} & \multicolumn{6}{c}{semi-supervised classification} \\
\cline{2-10}
 & \multicolumn{3}{c|}{K-Means} & \multicolumn{3}{c:}{KNN} & \multicolumn{3}{c}{Linear} \\
\cline{2-10}
\textbf{Model} & K=10 & K=20 & K=30 & 100 & 1000 & All & 100 & 1000 & All \\
\hline
ResNet VAE w. AF & 67.3 & 81.6 & 86.6 & 77.4 & 94.3 & 98.1 & 84.6 & 94.3 & 97.4 \\
VLAE & 68.1 & 74.0 & 79.1 & 75.7 & 90.0 & 95.6 & 86.4 & 93.7 & 96.1 \\
\hline
MAE: $\eta=1.0, \gamma=0.1$ & 82.7 & 92.3 & 93.0 & 86.6 & 95.5 & 97.8 & 91.1 & 96.3 & 98.3 \\
MAE: $\eta=0.5, \gamma=0.5$ & 84.7 & \textbf{92.6} & 93.2 & 86.3 & 96.3 & 98.0 & 90.6 & 96.1 & 98.1 \\
MAE: $\eta=1.0, \gamma=0.5$ & \textbf{91.2} & \textbf{92.6} & 93.6 & \textbf{86.7} & 95.9 & \textbf{98.2} & \textbf{91.5} & \textbf{96.4} & \textbf{98.4} \\
MAE: $\eta=2.0, \gamma=0.5$ & 78.2 & 92.0 & 92.8 & 85.5 & 96.4 & \textbf{98.2} & 90.7 & 96.0 & 98.0 \\
MAE: $\eta=2.0, \gamma=1.0$ & 83.1 & 92.3 & \textbf{94.3} & 86.2 & \textbf{96.6} & 98.1 & 90.0 & 95.7 & 98.0 \\
\hline
\end{tabular}
}
\end{table}

\subsubsection{Density Estimation}
We first evaluate MAE on density estimation performance.
Table~\ref{tab:density} provides the results of MAE with different settings of hyperparameters on MNIST, together with previous top systems for comparison.
Reported marginal negative log-likelihood (NLL) is evaluated with 4096 importance samples~\citep{burda2015importance}.
Our MAE achieves state-of-the-art performance on both the two datasets, exceeding all previous models and the re-implemented VLAE.
Note that our re-implementation of VLAE obtains better performance than the original one in \citet{chenvariational}, demonstrating the effectiveness of increasing decoder expressiveness by enlarging its receptive field.

\subsubsection{Representation Learning}\label{subsec:binary}
In order to evaluate the quality of the learned latent representations, we conduct three sets of experiments --- image reconstruction and generation, unsupervised clustering, and semi-supervised classification.

\begin{figure}[t]
\centering
\begin{minipage}[t]{0.99\textwidth}
\centering
\subfloat[MNIST reconstruction]{
\includegraphics[width=0.45\textwidth]{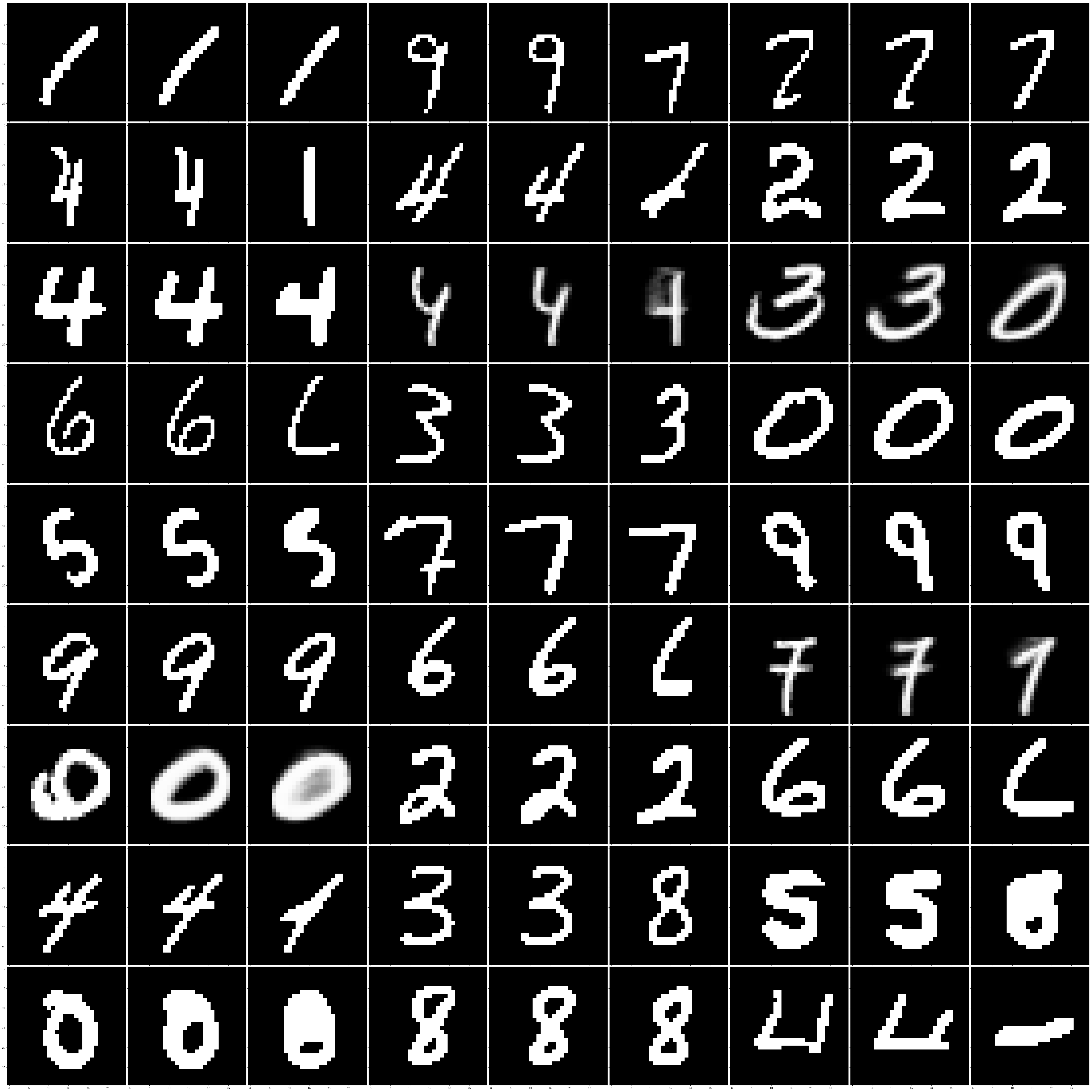}
\label{subfig:mnist_recon}
}
\subfloat[OMNIGLOT reconstruction]{
\includegraphics[width=0.45\textwidth]{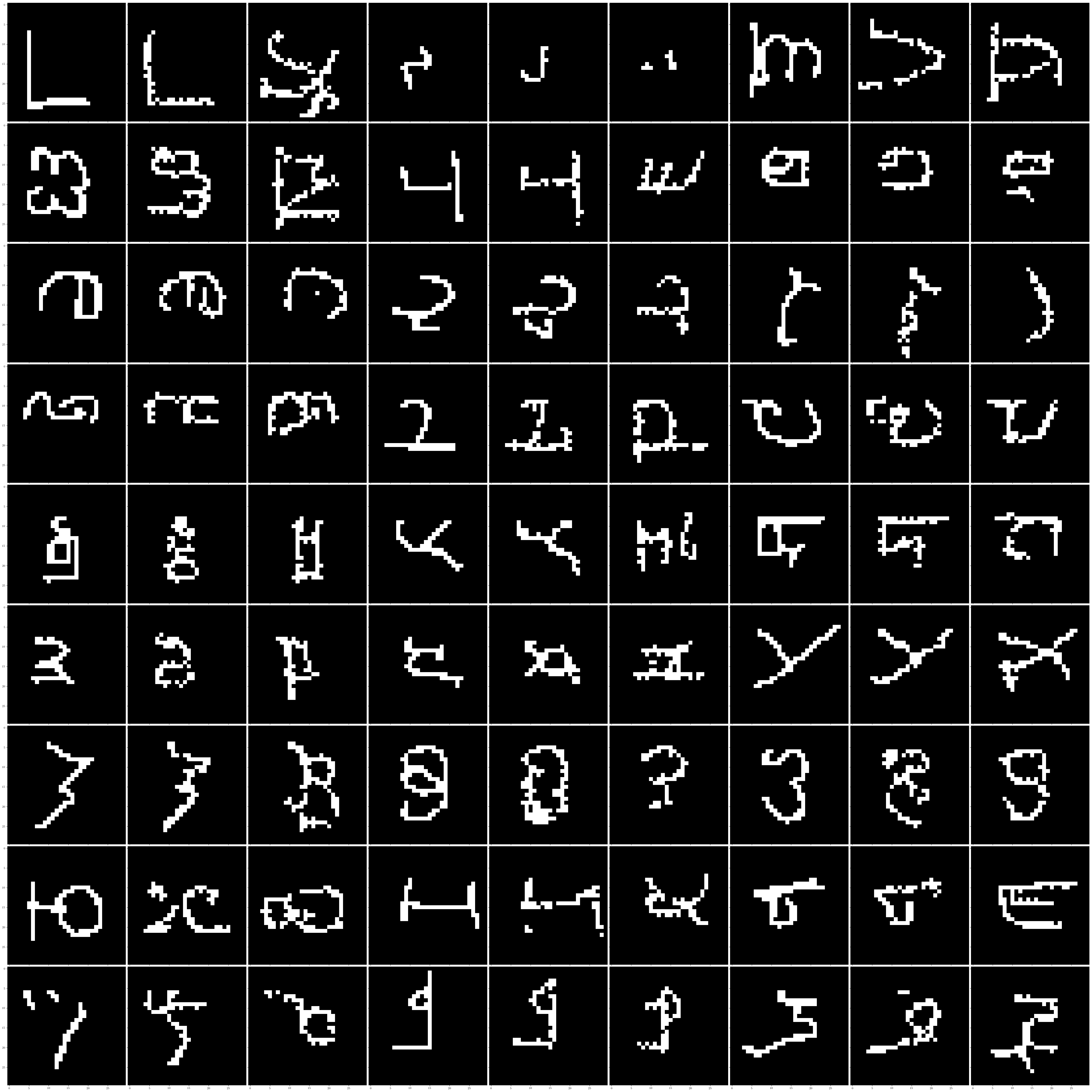}
\label{subfig:omni_recon}
}
\end{minipage}
\caption{Image reconstructions on MNIST and OMNIGLOT. 
Every three columns compose a set of reconstruction, original image is on the left, reconstructed image from MAE is in the middle, and reconstructed one from VLAE is on the right.}
\label{fig:binary_recon}
\end{figure}

\begin{figure}[t]
\centering
\begin{minipage}[t]{0.99\textwidth}
\centering
\subfloat[MNIST samples from MAE]{
\includegraphics[width=0.45\textwidth]{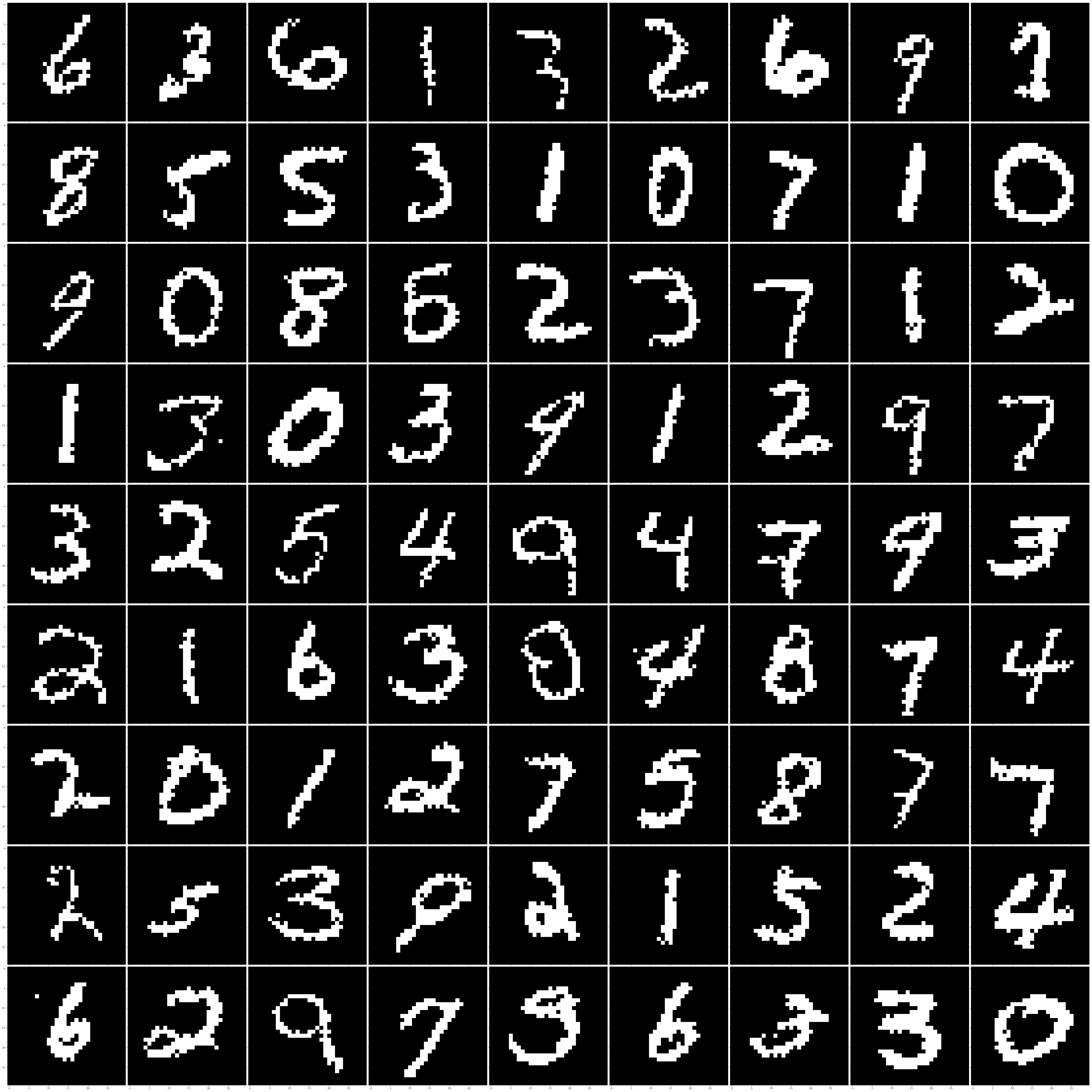}
\label{subfig:mnist_sample}
}
\subfloat[OMNIGLOT samples from MAE]{
\includegraphics[width=0.45\textwidth]{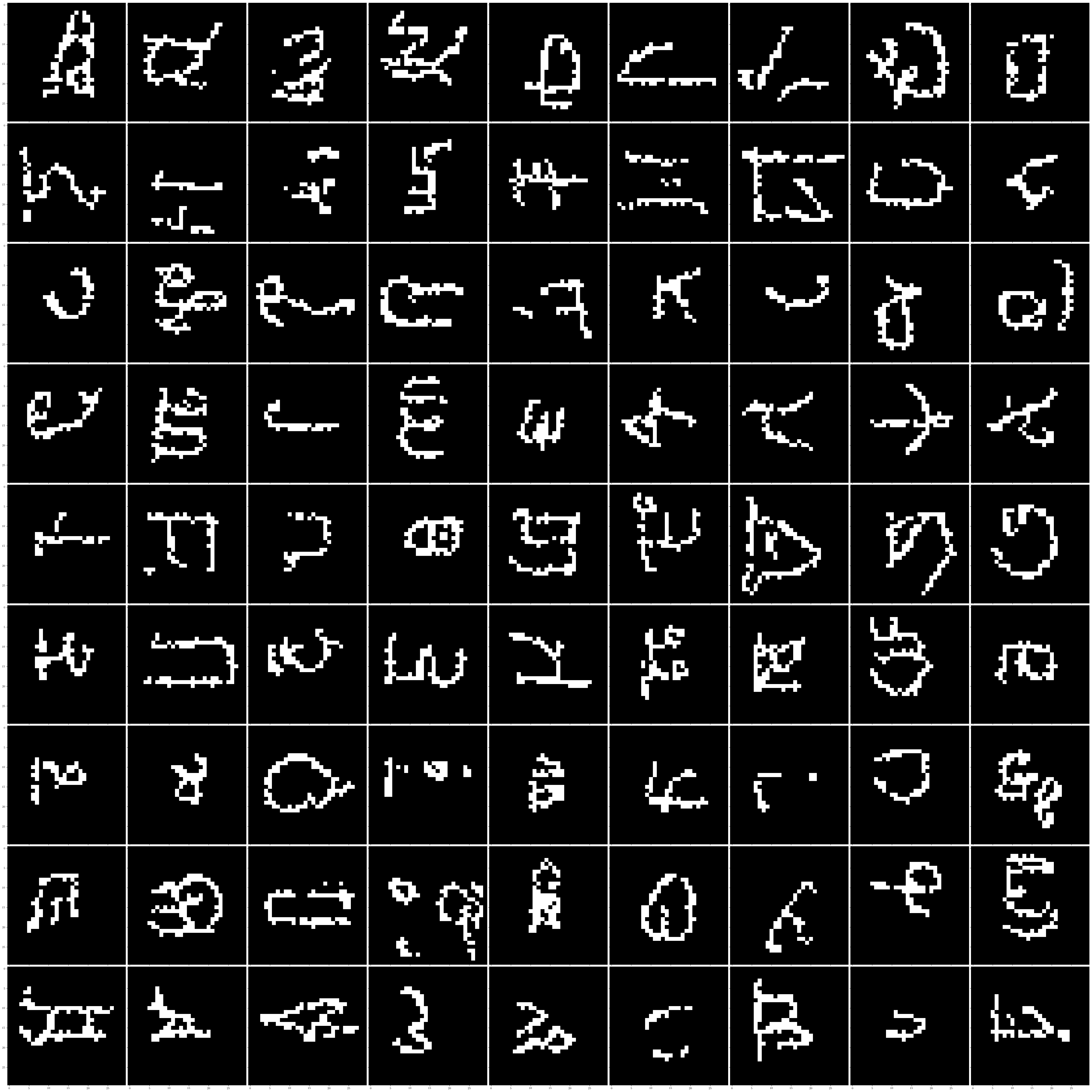}
\label{subfig:omni_sample}
}
\end{minipage}
\caption{Image samples on MNIST and OMNIGLOT from MAE.}
\label{fig:binary_sample}
\end{figure}

\paragraph{Image Reconstruction and Generation.}
The visualization of the of image reconstruction and generation on MNIST and OMNIGLOT is shown in Figure~\ref{fig:binary_recon} and Figure~\ref{fig:binary_sample}. 
For comparison, we also show the reconstructed images from VLAE. 
MAE achieves better reconstruction ability than VLAE, proving that the latent code from MAE encodes more information from data.

\paragraph{Unsupervised Clustering.}
As discussed above, good latent representations need to capture global structured information and disentangle the underlying causal factors, rather than just memorizing the data.
From this perspective, good image reconstruction results cannot guarantee good representations.
To further evaluate the quality of the learned representation from MAE, we conduct the experiments of unsupervised clustering on MNIST.
We perform K-Means clustering algorithm~\citep{hartigan1979algorithm} on the learned representations.
The class label of each cluster is assigned by finding the closest sample in the training data with the cluster head.
Evaluation of clustering accuracy is based on the assigned cluster labels.
We run three experiments with $K \in [10, 20, 30]$. 

Table~\ref{tab:classify} (left section) illustrates the clustering performance. 
To make a thorough comparison, we also re-implemented a VAE model with a factorized decoder $p_{\theta}(x|z) = \prod_{i} p_{\theta}(x_i|z)$ and AF prior, which has been proven to obtain remarkable reconstruction performance. 
The VAE model uses ResNet~\citep{he2016deep} as its encoder and decoder similar to \citet{chenvariational}.
From Table~\ref{tab:classify} we see that MAE significantly outperforms ResNet VAE and VLAE, especially when the number of clusters $K$ is small.
Interestingly, when $K$ keeps increasing, clustering accuracy of ResNet VAE increases rapidly, showing that in its latent space the data are split into small groups.

In addition, towards the affects of $\eta$ and $\gamma$ on learned representations, MAEs with larger $\eta$ obtain worse performance on $K=10$. 
The reason might be that large $\eta$ encourages the posteriors to diverse from each other, by splitting the data into small groups. 
Meanwhile, increasing $\gamma$ is effective to prevent the phenomenon,  showing that in practice considerations on the trade-off between space diversity and smoothness are needed.

\paragraph{Semi-supervised Classification.} 
For semi-supervised classification, we re-implemented the M1 model as described in \citet{kingma2014semi}.
To test quality of information encoded in the latent representations, we choose two simple classifiers with limited capacity --- K-nearest neighbor ($K=10$) and linear logistic regression.
For each classifier, we use different numbers of labeled data --- 100, 1000 and all the training data from MNIST.

From the results listed in Table~\ref{tab:classify} (right section), MAE obtains the best classification accuracy on all the settings.
Moreover, the improvements of MAE over ResNet VAE and VLAE are more significant when the number of labeled training data is small, further proving the meaningful representation learned from MAE.

\begin{table}[t]
\caption{Density estimation performance on CIFAR-10. Negative log-likelihood is evaluated with 512 importance samples.}
\label{tab:cifar10}
\centering
\begin{tabular}[t]{l|c}
\hline
\textbf{Model} & bits/dim \\
\hline
Deep GMMs~\citep{van2014factoring} & 4.00 \\
Real NVP~\citep{dinh2016density} & 3.49 \\
PixelCNN~\citep{oord2016pixel} & 3.14 \\
PixelRNN~\citep{oord2016pixel} & 3.00 \\
PixelCNN++~\citep{salimans2017pixelcnn++} & 2.92 \\
PixelSNAIL~\citep{chen2017pixelsnail} & \textbf{2.85} \\
\hline
Conv DRAW~\citep{gregor2016towards} & 3.50 \\
IAF VAE~\citep{kingma2016improved} & 3.11 \\
VLAE~\citep{chenvariational} & 2.95 \\
VLAE (re-impl) & 2.98 \\
\hdashline
MAE: $\eta=0.5, \gamma=1.5$ & \textbf{2.95} \\
MAE: $\eta=0.5, \gamma=2.0$ & 2.97 \\
MAE: $\eta=1.0, \gamma=2.0$ & 2.96 \\
\hline
\end{tabular}
\end{table}

\subsection{Natural Images}\label{subsec:natural}
In addition to binary image datasets, we also applied MAE to CIFAR-10 dataset~\citep{krizhevsky2009learning} of natural images.
The VLAE with DenseNet~\citep{huang2017densely} encoder and PixelCNN++~\citep{salimans2017pixelcnn++} decoder described in \citet{chenvariational} is used as the neural architecture of MAE.
To ensure that the decoder is sufficiently expressive, the decoder PixelCNN has 5 blocks of 96 feature maps and 7x4 receptive field.
Hence the PixelCNN decoder we used is both deeper and wider than that used in \citet{chenvariational}.

\subsection{Density Estimation}
Density estimation performance on CIFAR-10 of MAEs with different hyperparameters is provided in Table~\ref{tab:cifar10}, compared with the top-performing likelihood-based unconditional generative models (first section) and variationally trained latent-variable models (second section).
MAE models obtain improvement over the VLAE re-implemented by us, and slightly fall behind the original one in \citet{chenvariational}.
Compared with PixelSNAIL~\citep{chen2017pixelsnail}, the state-of-the-art auto-regressive generative model, the performance of MAE models is around 0.11 bits/dim worse.
Further improving the density estimation performance of MAEs on natural images has been left to future work.

\begin{figure}[t]
\centering
\begin{minipage}[t]{0.99\textwidth}
\centering
\subfloat[CIFAR-10 reconstruction]{
\includegraphics[width=0.47\textwidth]{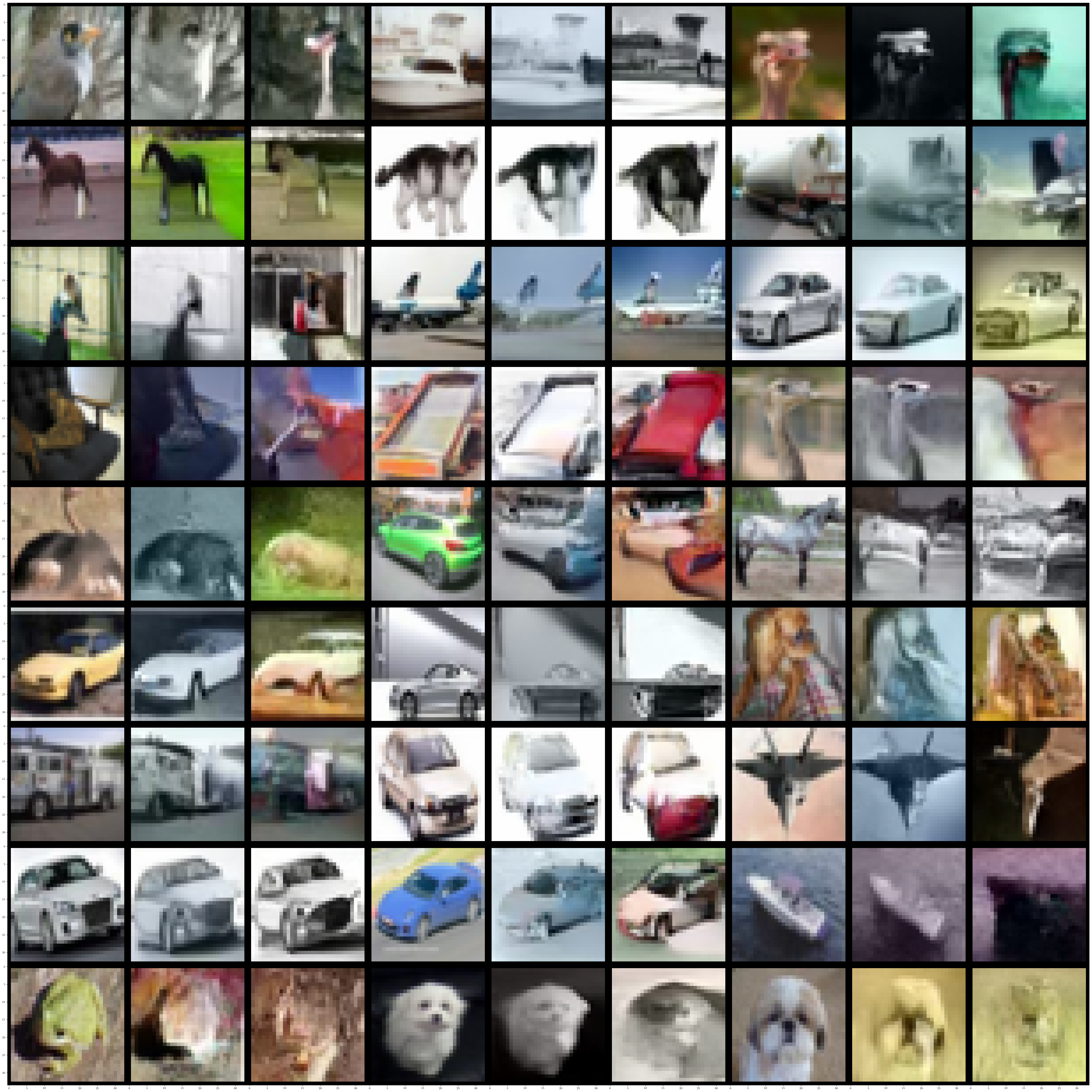}
\label{subfig:cifar_recon}
}
\subfloat[CIFAR-10 samples from MAE]{
\includegraphics[width=0.47\textwidth]{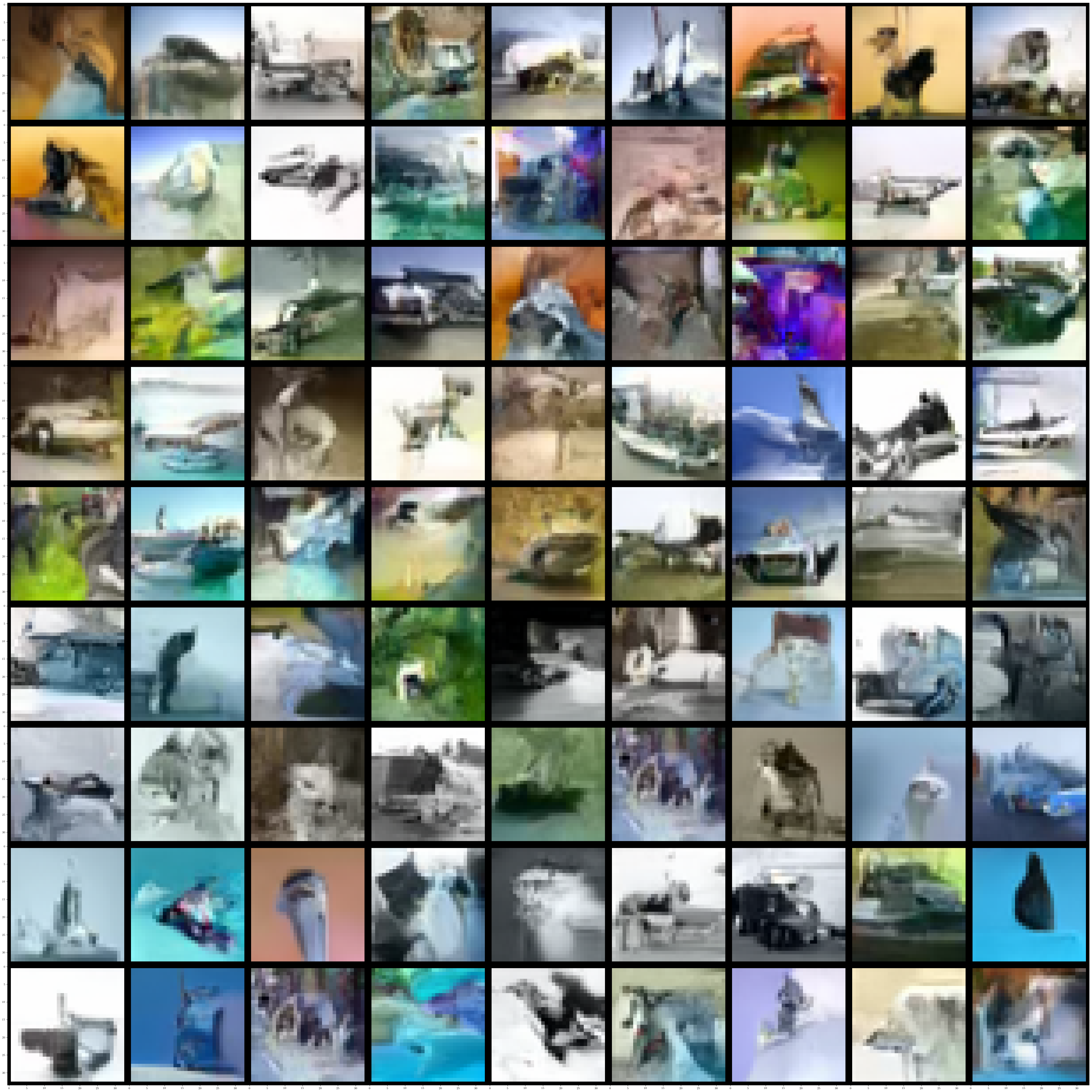}
\label{subfig:cifar_sample}
}
\end{minipage}
\caption{Image reconstructions and samples on CIFAR-10. 
For reconstruction, original image is on the left, reconstructed image from MAE is in the middle, and VLAE is on the right.}
\label{fig:cifar10}
\end{figure}

\subsection{Image Reconstruction and Generation}
We also investigate learning informative representations on CIFAR-10 dataset. 
The visualization of image reconstruction and generation is shown in Figure~\ref{fig:cifar10}, together with VLAE for comparison.
It is interesting to note that MAE tends to preserve rather detailed shape information than VLAE, whereas the color information, particularly the color for background, is partially omitted.
One reasonable explanation, as discussed in \citet{chenvariational}, is that color is predictable locally.
This serves as one example showing that MAEs can capture global structured information from data, omitting common correlations.
Image samples from MAE are shown in Figure~\ref{subfig:cifar_sample}.

\section{Conclusion}
In this paper, we proposed a mutual posterior-divergence regularization for VAEs, which controls the geometry of the latent space during training. 
By connecting the mutual posterior diversity with the mutual information, we have formally studied the theoretical properties of the proposed MAEs.
Experiments on three benchmark datasets of images show the capability of MAEs on both density estimation and representation learning, with state-of-the-art or comparable likelihood, and superior performance on image reconstruction, unsupervised clustering and semi-supervised classification against previous top-performing models.

One potential direction for future work is to extend MAE to other forms of data, in particular text on which VAEs suffer a more serious KL-varnishing problem.
Another exciting direction is to formally study the properties of the standard deviation of the mutual posterior KL-divergence used to measure smoothness, hence providing further justification of the proposed regularizer, or even introducing alternatives to further improve performances.

\section*{Acknowledgements}
The authors thank Zihang Dai, Junxian He, Di Wang and Zhengzhong Liu for their helpful discussions.
This research was supported in part by DARPA grant FA8750-18-2-0018 funded under the AIDA program. 
Any opinions, findings, and conclusions or recommendations expressed in this material are those of the authors and do not necessarily reflect the views of DARPA.

\bibliography{iclr2019_conference}

\begin{thebibliography}{47}
\providecommand{\natexlab}[1]{#1}
\providecommand{\url}[1]{\texttt{#1}}
\expandafter\ifx\csname urlstyle\endcsname\relax
  \providecommand{\doi}[1]{doi: #1}\else
  \providecommand{\doi}{doi: \begingroup \urlstyle{rm}\Url}\fi

\bibitem[Bengio et~al.(2013)Bengio, Courville, and
  Vincent]{bengio2013representation}
Yoshua Bengio, Aaron Courville, and Pascal Vincent.
\newblock Representation learning: A review and new perspectives.
\newblock \emph{IEEE transactions on pattern analysis and machine
  intelligence}, 35\penalty0 (8):\penalty0 1798--1828, 2013.

\bibitem[Bowman et~al.(2015)Bowman, Vilnis, Vinyals, Dai, Jozefowicz, and
  Bengio]{bowman2015generating}
Samuel~R Bowman, Luke Vilnis, Oriol Vinyals, Andrew~M Dai, Rafal Jozefowicz,
  and Samy Bengio.
\newblock Generating sentences from a continuous space.
\newblock \emph{arXiv preprint arXiv:1511.06349}, 2015.

\bibitem[Burda et~al.(2015)Burda, Grosse, and
  Salakhutdinov]{burda2015importance}
Yuri Burda, Roger Grosse, and Ruslan Salakhutdinov.
\newblock Importance weighted autoencoders.
\newblock \emph{arXiv preprint arXiv:1509.00519}, 2015.

\bibitem[Chen et~al.(2016)Chen, Duan, Houthooft, Schulman, Sutskever, and
  Abbeel]{chen2016infogan}
Xi~Chen, Yan Duan, Rein Houthooft, John Schulman, Ilya Sutskever, and Pieter
  Abbeel.
\newblock Infogan: Interpretable representation learning by information
  maximizing generative adversarial nets.
\newblock In \emph{Advances in neural information processing systems}, pp.\
  2172--2180, 2016.

\bibitem[Chen et~al.(2017{\natexlab{a}})Chen, Kingma, Salimans, Duan, Dhariwal,
  Schulman, Sutskever, and Abbeel]{chenvariational}
Xi~Chen, Diederik~P Kingma, Tim Salimans, Yan Duan, Prafulla Dhariwal, John
  Schulman, Ilya Sutskever, and Pieter Abbeel.
\newblock Variational lossy autoencoder.
\newblock In \emph{Proceedings of the 5th International Conference on Learning
  Representations (ICLR-2017)}, Toulon, France, April 2017{\natexlab{a}}.

\bibitem[Chen et~al.(2017{\natexlab{b}})Chen, Mishra, Rohaninejad, and
  Abbeel]{chen2017pixelsnail}
Xi~Chen, Nikhil Mishra, Mostafa Rohaninejad, and Pieter Abbeel.
\newblock Pixelsnail: An improved autoregressive generative model.
\newblock \emph{arXiv preprint arXiv:1712.09763}, 2017{\natexlab{b}}.

\bibitem[Dinh et~al.(2016)Dinh, Sohl-Dickstein, and Bengio]{dinh2016density}
Laurent Dinh, Jascha Sohl-Dickstein, and Samy Bengio.
\newblock Density estimation using real nvp.
\newblock \emph{arXiv preprint arXiv:1605.08803}, 2016.

\bibitem[Dziugaite et~al.(2015)Dziugaite, Roy, and
  Ghahramani]{dziugaite2015training}
Gintare~Karolina Dziugaite, Daniel~M Roy, and Zoubin Ghahramani.
\newblock Training generative neural networks via maximum mean discrepancy
  optimization.
\newblock \emph{arXiv preprint arXiv:1505.03906}, 2015.

\bibitem[Esmaeili et~al.(2018)Esmaeili, Wu, Jain, Bozkurt, Siddharth, Paige,
  Brooks, Dy, and van~de Meent]{esmaeili2018structured}
Babak Esmaeili, Hao Wu, Sarthak Jain, Alican Bozkurt, N~Siddharth, Brooks
  Paige, Dana~H Brooks, Jennifer Dy, and Jan-Willem van~de Meent.
\newblock Structured disentangled representations.
\newblock \emph{stat}, 1050:\penalty0 29, 2018.

\bibitem[Ganchev et~al.(2010)Ganchev, Gillenwater, Taskar,
  et~al.]{ganchev2010posterior}
Kuzman Ganchev, Jennifer Gillenwater, Ben Taskar, et~al.
\newblock Posterior regularization for structured latent variable models.
\newblock \emph{Journal of Machine Learning Research}, 11\penalty0
  (Jul):\penalty0 2001--2049, 2010.

\bibitem[Germain et~al.(2015)Germain, Gregor, Murray, and
  Larochelle]{germain2015made}
Mathieu Germain, Karol Gregor, Iain Murray, and Hugo Larochelle.
\newblock Made: Masked autoencoder for distribution estimation.
\newblock In \emph{International Conference on Machine Learning}, pp.\
  881--889, 2015.

\bibitem[Goodfellow et~al.(2014)Goodfellow, Pouget-Abadie, Mirza, Xu,
  Warde-Farley, Ozair, Courville, and Bengio]{goodfellow2014generative}
Ian Goodfellow, Jean Pouget-Abadie, Mehdi Mirza, Bing Xu, David Warde-Farley,
  Sherjil Ozair, Aaron Courville, and Yoshua Bengio.
\newblock Generative adversarial nets.
\newblock In \emph{Advances in neural information processing systems
  (NIPS-2014)}, pp.\  2672--2680, 2014.

\bibitem[Gregor et~al.(2016)Gregor, Besse, Rezende, Danihelka, and
  Wierstra]{gregor2016towards}
Karol Gregor, Frederic Besse, Danilo~Jimenez Rezende, Ivo Danihelka, and Daan
  Wierstra.
\newblock Towards conceptual compression.
\newblock In \emph{Advances In Neural Information Processing Systems}, pp.\
  3549--3557, 2016.

\bibitem[Gretton et~al.(2007)Gretton, Borgwardt, Rasch, Sch{\"o}lkopf, and
  Smola]{gretton2007kernel}
Arthur Gretton, Karsten~M Borgwardt, Malte Rasch, Bernhard Sch{\"o}lkopf, and
  Alex~J Smola.
\newblock A kernel method for the two-sample-problem.
\newblock In \emph{Advances in neural information processing systems}, pp.\
  513--520, 2007.

\bibitem[Hartigan \& Wong(1979)Hartigan and Wong]{hartigan1979algorithm}
John~A Hartigan and Manchek~A Wong.
\newblock Algorithm as 136: A k-means clustering algorithm.
\newblock \emph{Journal of the Royal Statistical Society. Series C (Applied
  Statistics)}, 28\penalty0 (1):\penalty0 100--108, 1979.

\bibitem[He et~al.(2016)He, Zhang, Ren, and Sun]{he2016deep}
Kaiming He, Xiangyu Zhang, Shaoqing Ren, and Jian Sun.
\newblock Deep residual learning for image recognition.
\newblock In \emph{Proceedings of the IEEE conference on computer vision and
  pattern recognition}, pp.\  770--778, 2016.

\bibitem[Hinton et~al.(1995)Hinton, Dayan, Frey, and Neal]{hinton1995wake}
Geoffrey~E Hinton, Peter Dayan, Brendan~J Frey, and Radford~M Neal.
\newblock The" wake-sleep" algorithm for unsupervised neural networks.
\newblock \emph{Science}, 268\penalty0 (5214):\penalty0 1158--1161, 1995.

\bibitem[Hjelm et~al.(2018)Hjelm, Fedorov, Lavoie-Marchildon, Grewal,
  Trischler, and Bengio]{hjelm2018learning}
R~Devon Hjelm, Alex Fedorov, Samuel Lavoie-Marchildon, Karan Grewal, Adam
  Trischler, and Yoshua Bengio.
\newblock Learning deep representations by mutual information estimation and
  maximization.
\newblock \emph{arXiv preprint arXiv:1808.06670}, 2018.

\bibitem[Huang et~al.(2017)Huang, Liu, Van Der~Maaten, and
  Weinberger]{huang2017densely}
Gao Huang, Zhuang Liu, Laurens Van Der~Maaten, and Kilian~Q Weinberger.
\newblock Densely connected convolutional networks.
\newblock In \emph{CVPR}, volume~1, pp.\ ~3, 2017.

\bibitem[JLB(2015)]{jlb2015adam}
Diederik P~Kingma JLB.
\newblock Adam: A method for stochastic optimization.
\newblock \emph{Proc. of ICLR}, 2015.

\bibitem[Kim et~al.(2018)Kim, Wiseman, Miller, Sontag, and Rush]{kim2018semi}
Yoon Kim, Sam Wiseman, Andrew~C Miller, David Sontag, and Alexander~M Rush.
\newblock Semi-amortized variational autoencoders.
\newblock \emph{arXiv preprint arXiv:1802.02550}, 2018.

\bibitem[Kingma \& Welling(2014)Kingma and Welling]{kingma2014auto}
Diederik~P Kingma and Max Welling.
\newblock Auto-encoding variational bayes.
\newblock In \emph{Proceedings of the 2th International Conference on Learning
  Representations (ICLR-2014)}, Banff, Canada, April 2014.

\bibitem[Kingma et~al.(2014)Kingma, Mohamed, Rezende, and
  Welling]{kingma2014semi}
Diederik~P Kingma, Shakir Mohamed, Danilo~Jimenez Rezende, and Max Welling.
\newblock Semi-supervised learning with deep generative models.
\newblock In \emph{Advances in Neural Information Processing Systems}, pp.\
  3581--3589, 2014.

\bibitem[Kingma et~al.(2016)Kingma, Salimans, Jozefowicz, Chen, Sutskever, and
  Welling]{kingma2016improved}
Diederik~P Kingma, Tim Salimans, Rafal Jozefowicz, Xi~Chen, Ilya Sutskever, and
  Max Welling.
\newblock Improved variational inference with inverse autoregressive flow.
\newblock In \emph{Advances in Neural Information Processing Systems}, pp.\
  4743--4751, 2016.

\bibitem[Krause et~al.(2010)Krause, Perona, and
  Gomes]{krause2010discriminative}
Andreas Krause, Pietro Perona, and Ryan~G Gomes.
\newblock Discriminative clustering by regularized information maximization.
\newblock In \emph{Advances in neural information processing systems}, pp.\
  775--783, 2010.

\bibitem[Krizhevsky \& Hinton(2009)Krizhevsky and
  Hinton]{krizhevsky2009learning}
Alex Krizhevsky and Geoffrey Hinton.
\newblock Learning multiple layers of features from tiny images.
\newblock Technical report, Citeseer, 2009.

\bibitem[Lake et~al.(2013)Lake, Salakhutdinov, and Tenenbaum]{lake2013one}
Brenden~M Lake, Ruslan~R Salakhutdinov, and Josh Tenenbaum.
\newblock One-shot learning by inverting a compositional causal process.
\newblock In \emph{Advances in neural information processing systems}, pp.\
  2526--2534, 2013.

\bibitem[Larochelle \& Murray(2011)Larochelle and Murray]{larochelle2011neural}
Hugo Larochelle and Iain Murray.
\newblock The neural autoregressive distribution estimator.
\newblock In \emph{Proceedings of the Fourteenth International Conference on
  Artificial Intelligence and Statistics (AISTATS-2011}, pp.\  29--37, 2011.

\bibitem[LeCun et~al.(1998)LeCun, Bottou, Bengio, and
  Haffner]{lecun1998gradient}
Yann LeCun, L{\'e}on Bottou, Yoshua Bengio, and Patrick Haffner.
\newblock Gradient-based learning applied to document recognition.
\newblock \emph{Proceedings of the IEEE}, 86\penalty0 (11):\penalty0
  2278--2324, 1998.

\bibitem[Li et~al.(2015)Li, Swersky, and Zemel]{li2015generative}
Yujia Li, Kevin Swersky, and Rich Zemel.
\newblock Generative moment matching networks.
\newblock In \emph{Proceedings of International Conference on Machine Learning
  (ICML-2015)}, pp.\  1718--1727, 2015.

\bibitem[Maaten \& Hinton(2008)Maaten and Hinton]{maaten2008visualizing}
Laurens van~der Maaten and Geoffrey Hinton.
\newblock Visualizing data using t-sne.
\newblock \emph{Journal of machine learning research}, 9\penalty0
  (Nov):\penalty0 2579--2605, 2008.

\bibitem[Makhzani et~al.(2015)Makhzani, Shlens, Jaitly, Goodfellow, and
  Frey]{makhzani2015adversarial}
Alireza Makhzani, Jonathon Shlens, Navdeep Jaitly, Ian Goodfellow, and Brendan
  Frey.
\newblock Adversarial autoencoders.
\newblock \emph{arXiv preprint arXiv:1511.05644}, 2015.

\bibitem[Oord et~al.(2016)Oord, Kalchbrenner, and Kavukcuoglu]{oord2016pixel}
Aaron van~den Oord, Nal Kalchbrenner, and Koray Kavukcuoglu.
\newblock Pixel recurrent neural networks.
\newblock In \emph{Proceedings of International Conference on Machine Learning
  (ICML-2016)}, 2016.

\bibitem[Polyak \& Juditsky(1992)Polyak and Juditsky]{polyak1992acceleration}
Boris~T Polyak and Anatoli~B Juditsky.
\newblock Acceleration of stochastic approximation by averaging.
\newblock \emph{SIAM Journal on Control and Optimization}, 30\penalty0
  (4):\penalty0 838--855, 1992.

\bibitem[Rezende \& Mohamed(2015)Rezende and Mohamed]{rezende2015variational}
Danilo~Jimenez Rezende and Shakir Mohamed.
\newblock Variational inference with normalizing flows.
\newblock \emph{arXiv preprint arXiv:1505.05770}, 2015.

\bibitem[Rezende et~al.(2014)Rezende, Mohamed, and Wierstra]{rezende14}
Danilo~Jimenez Rezende, Shakir Mohamed, and Daan Wierstra.
\newblock Stochastic backpropagation and approximate inference in deep
  generative models.
\newblock In \emph{Proceedings of the 31st International Conference on Machine
  Learning (ICML-2014)}, pp.\  1278--1286, Bejing, China, 22--24 Jun 2014.

\bibitem[Rolfe(2016)]{rolfe2016discrete}
Jason~Tyler Rolfe.
\newblock Discrete variational autoencoders.
\newblock \emph{arXiv preprint arXiv:1609.02200}, 2016.

\bibitem[Salimans et~al.(2017)Salimans, Karpathy, Chen, Kingma, and
  Bulatov]{salimans2017pixelcnn++}
Tim Salimans, Andrej Karpathy, Xi~Chen, Diederik~P Kingma, and Yaroslav
  Bulatov.
\newblock Pixelcnn++: A pixelcnn implementation with discretized logistic
  mixture likelihood and other modifications.
\newblock In \emph{International Conference on Learning Representations
  (ICLR)}, 2017.

\bibitem[Serban et~al.(2017)Serban, Sordoni, Lowe, Charlin, Pineau, Courville,
  and Bengio]{serban2017hierarchical}
Iulian~Vlad Serban, Alessandro Sordoni, Ryan Lowe, Laurent Charlin, Joelle
  Pineau, Aaron~C Courville, and Yoshua Bengio.
\newblock A hierarchical latent variable encoder-decoder model for generating
  dialogues.
\newblock In \emph{AAAI}, pp.\  3295--3301, 2017.

\bibitem[S{\o}nderby et~al.(2016{\natexlab{a}})S{\o}nderby, Raiko, Maal{\o}e,
  S{\o}nderby, and Winther]{sonderby2016ladder}
Casper~Kaae S{\o}nderby, Tapani Raiko, Lars Maal{\o}e, S{\o}ren~Kaae
  S{\o}nderby, and Ole Winther.
\newblock Ladder variational autoencoders.
\newblock In \emph{Advances in neural information processing systems}, pp.\
  3738--3746, 2016{\natexlab{a}}.

\bibitem[S{\o}nderby et~al.(2016{\natexlab{b}})S{\o}nderby, Raiko, Maal{\o}e,
  S{\o}nderby, and Winther]{sonderby2016train}
Casper~Kaae S{\o}nderby, Tapani Raiko, Lars Maal{\o}e, S{\o}ren~Kaae
  S{\o}nderby, and Ole Winther.
\newblock How to train deep variational autoencoders and probabilistic ladder
  networks.
\newblock \emph{arXiv preprint arXiv:1602.02282}, 2016{\natexlab{b}}.

\bibitem[Tomczak \& Welling(2018)Tomczak and Welling]{tomczak2018vae}
Jakub Tomczak and Max Welling.
\newblock Vae with a vampprior.
\newblock In \emph{International Conference on Artificial Intelligence and
  Statistics}, pp.\  1214--1223, 2018.

\bibitem[Van~den Oord \& Schrauwen(2014)Van~den Oord and
  Schrauwen]{van2014factoring}
Aaron Van~den Oord and Benjamin Schrauwen.
\newblock Factoring variations in natural images with deep gaussian mixture
  models.
\newblock In \emph{Advances in Neural Information Processing Systems}, pp.\
  3518--3526, 2014.

\bibitem[Wainwright et~al.(2008)Wainwright, Jordan,
  et~al.]{wainwright2008graphical}
Martin~J Wainwright, Michael~I Jordan, et~al.
\newblock Graphical models, exponential families, and variational inference.
\newblock \emph{Foundations and Trends{\textregistered} in Machine Learning},
  1\penalty0 (1--2):\penalty0 1--305, 2008.

\bibitem[Xie et~al.(2015)Xie, Deng, and Xing]{xie2015latent}
Pengtao Xie, Yuntian Deng, and Eric Xing.
\newblock Latent variable modeling with diversity-inducing mutual angular
  regularization.
\newblock \emph{arXiv preprint arXiv:1512.07336}, 2015.

\bibitem[Yang et~al.(2017)Yang, Hu, Salakhutdinov, and
  Berg-Kirkpatrick]{yang2017improved}
Zichao Yang, Zhiting Hu, Ruslan Salakhutdinov, and Taylor Berg-Kirkpatrick.
\newblock Improved variational autoencoders for text modeling using dilated
  convolutions.
\newblock In \emph{Proceedings of International Conference on Machine Learning
  (ICML-2017)}, 2017.

\bibitem[Zhao et~al.(2017)Zhao, Song, and Ermon]{zhao2017infovae}
Shengjia Zhao, Jiaming Song, and Stefano Ermon.
\newblock Infovae: Information maximizing variational autoencoders.
\newblock \emph{arXiv preprint arXiv:1706.02262}, 2017.

\end{thebibliography}
\bibliographystyle{iclr2019_conference}

\newpage
\section*{Appendix: MAE: Mutual Posterior-Divergence Regularization for Variational AutoEncoders}
\appendix
\setcounter{equation}{0}
\section{Proof of Theorem 1}\label{appendix:proof:thm1}
\begin{proof}
\begin{displaymath}
{\arraycolsep=2pt\def\arraystretch{1.7}
\begin{array}{rcl}
MPD & = & \mathrm{E}_{X_1, X_2\sim P(X)} [\mathrm{KL}(q_{\phi}(Z|X_1) || q_{\phi}(Z|X_2))] \\
 & = & \mathrm{E}_{X_1, X_2\sim P(X)} \left[ \mathrm{H}(q_{\phi}(Z|X_1), q_{\phi}(Z|X_2)) - \mathrm{H}(q_{\phi}(Z|X_1)) \right]
\end{array}
}
\end{displaymath}
where $\mathrm{H}(\cdot)$ denotes the entropy.
Then,
\begin{displaymath}
\mathrm{E}_{X_1, X_2\sim P(X)} \left[\mathrm{H}(q_{\phi}(Z|X_1)) \right] = \mathrm{E}_{P(X)} \left[\mathrm{H}(q_{\phi}(Z|X)) \right]
\end{displaymath}
and
\begin{displaymath}
{\arraycolsep=2pt\def\arraystretch{1.7}
\begin{array}{rl}
& \mathrm{E}_{X_1, X_2\sim P(X)} \left[ \mathrm{H}(q_{\phi}(Z|X_1), q_{\phi}(Z|X_2))\right] \\
= & \mathrm{E}_{X_1, X_2\sim P(X)} \left[ - \int_{\mathcal{Z}} q_{\phi}(z|x_1) \log q_{\phi}(z|x_2) dz \right] \\
= & \mathrm{E}_{P(X_2)} \left[ - \int\!\!\int p(x_1) q_{\phi}(z|x_1) \log q_{\phi}(z|x_2) dz\,dx_1 \right] \\
= & \mathrm{E}_{P(X_2)} \left[ - \int \left(\int p(x_1) q_{\phi}(z|x_1) dx_1\right) \log q_{\phi}(z|x_2) dz \right] \\
= & \mathrm{E}_{P(X_2)} \left[ - \int q_{\phi}(z) \log q_{\phi}(z|x_2) dz \right] \\
= & \mathrm{E}_{P(X)} \left[ \mathrm{H}(q_{\phi}(Z), q_{\phi}(Z|X)) \right]
\end{array}
}
\end{displaymath}
So we have, 
\begin{displaymath}
{\arraycolsep=2pt\def\arraystretch{1.7}
\begin{array}{rcl}
MPD & = & \mathrm{E}_{P(X)} \left[ \mathrm{H}(q_{\phi}(Z), q_{\phi}(Z|X)) - \mathrm{H}(q_{\phi}(Z|X)) \right] \\
 & = & \mathrm{E}_{P(X)} \left[ \mathrm{H}(q_{\phi}(Z), q_{\phi}(Z|X)) - \mathrm{H}(q_{\phi}(Z)) + \mathrm{H}(q_{\phi}(Z)) - \mathrm{H}(q_{\phi}(Z|X)) \right] \\
 & = & \mathrm{E}_{P(X)} [\mathrm{KL} (q_{\phi}(z|x) || q_{\phi}(z)) + \mathrm{KL} (q_{\phi}(z) || q_{\phi}(z|x))]
\end{array}
}
\end{displaymath}
\end{proof}

\section{Detailed Description of Experiments}\label{appendix:experiment}
\subsection{Experiments for Binary Images}\label{appendix:binary}
\subsubsection{Neural Network Architectures and Training}
The neural network architectures, including most of the hyperparameters, are the same as those in \citet{chenvariational}.
The only difference in network architecture is the filter size of the PixelCNN decoder, which has been described in \S\ref{sub:experiment}.
For ResNet VAE with AF, we use the same ResNet encoder but a symmetric ResNet architecture for decoder.
For encoder, we only use one stochastic layer with 32 dimensions.

In term of training, we use Adam optimizer~\citep{jlb2015adam} with learning rate 0.001, instead of Adamax used in \citet{chenvariational}.
0.01 nats/data-dim free bits was used in all the experiments.
In order to get a relatively accurate approximation of $\mathcal{L}_{diverse}$ and $\mathcal{L}_{smooth}$, we used a much larger batch size 100 in our experiments.
Polyak averaging~\citep{polyak1992acceleration} was used to compute the final parameters, with $\alpha=0.999$.

\begin{table}
\caption{Density estimation results on dynamically binarized MNIST.
RE and KL correspond to the reconstruction error and the KL term in ELBO. MPD is the mutual posterior diversity in \eqref{eq:mpd}, and STD is the corresponding standard deviation in \eqref{eq:std}.
}
\label{tab:mnist}
\centering
\begin{tabular}[t]{l|cccccc}
\hline
\textbf{Model} & RE & KL & MPD & STD & $\mathcal{L}_{elbo}$ & NLL \\
\hline
ResNet VAE with AF & 56.04 & 25.38 & 1,193.18 & 630.90 & 81.42 & 79.28 \\
VLAE (w.o free bits) & 71.74 & 7.07 & 109.31 & 66.52 & 78.81 & 78.45 \\
VLAE (w. free bits)  & 69.60 & 9.02 & 132.00 & 66.30 & 78.62 & 78.26 \\
\hline
MAE ($\eta=0.5, \gamma=0.1$) & 69.58 & 9.57 & 99.65 & 22.03	& 79.15 & 78.04 \\
MAE ($\eta=1.0, \gamma=0.1$) & 67.95 & 11.38 & 124.80 & 26.40 & 79.33 & 78.02 \\
MAE ($\eta=2.0, \gamma=0.1$) & 66.83 & 12.67 & 148.84 & 29.88 & 79.50 & 78.02 \\
\hdashline
MAE ($\eta=0.5, \gamma=0.5$) & 68.44 & 10.44 & 55.09 & 8.93 & 78.88 & 78.00 \\
MAE ($\eta=1.0, \gamma=0.5$) & 67.40 & 11.54 & 79.71 & 12.69 & 78.94 & 77.98 \\
MAE ($\eta=2.0, \gamma=0.5$) & 66.54 & 12.67 & 103.30 & 16.32 & 79.04 & 77.99 \\
\hdashline
MAE ($\eta=0.5, \gamma=1.0$) & 71.08 & 8.41 & 30.64 & 4.78 & 79.49 & 78.36 \\
MAE ($\eta=1.0, \gamma=1.0$) & 69.34 & 10.19 & 55.59 & 8.51 & 79.53 & 78.15 \\
MAE ($\eta=2.0, \gamma=1.0$) & 68.03 & 11.65 & 80.87 & 12.19 & 79.68 & 78.06 \\
\hline
\end{tabular}
\end{table}

\subsubsection{Detailed Results on Density Estimation}
Table~\ref{tab:mnist} shows the detailed results of density estimation on MNIST. 
We see that increasing $\eta$ always achieving more informative latent $Z$, but the NLL not always becomes better.
It illustrates the hypothesis that good representations should encode global structured information in the data, rather than local dependencies.
It is interesting to see that, the effect of $\gamma$ on the latent $Z$ is inconsistent --- increasing $\gamma$ from 0.1 to 0.5 leads to more informative $Z$ (larger KL) and better NLL, but too large $\gamma$ (1.0) prevent the latent $Z$ to learn more information from data (smaller KL), resulting worse NLL.
Hence, in practice considerations on the trade-off between diversity and smoothness of the latent space are needed.

\begin{figure}[t]
\begin{minipage}[t]{0.99\textwidth}
\flushright
\subfloat{
\includegraphics[width=0.96\textwidth]{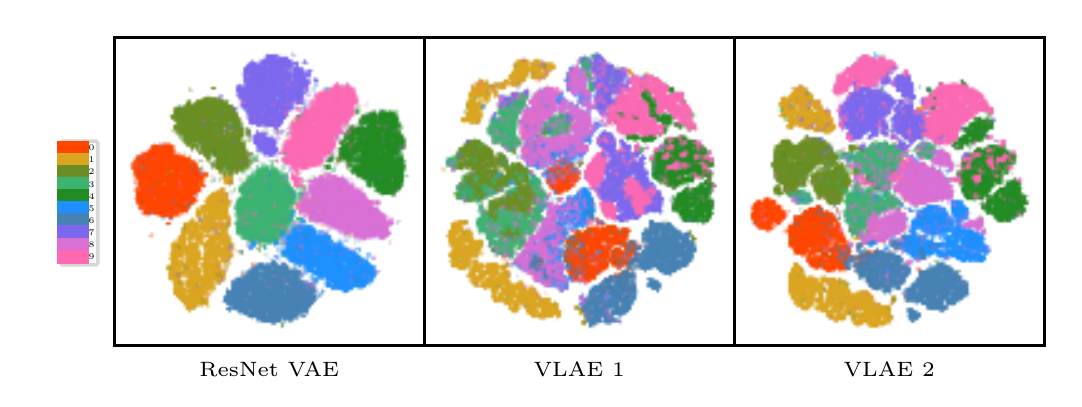}
}
\\
\subfloat{
\includegraphics[width=0.98\textwidth]{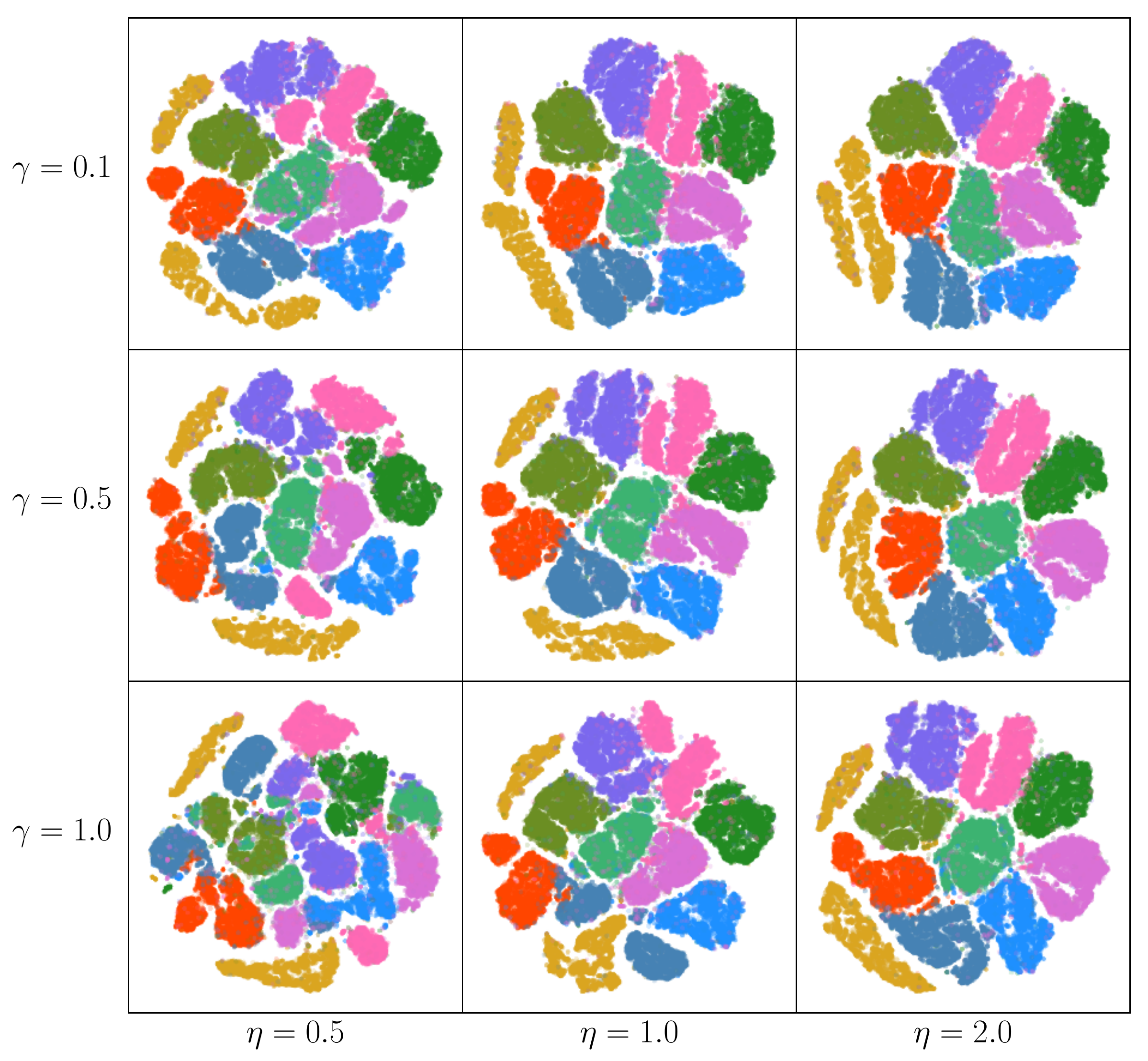}
}
\end{minipage}
\caption{Visualization of latent space via t-SNE.}
\label{fig:tsne}
\end{figure}

\subsubsection{Detailed Results on Semi-supervised Classification with SVMs}
Table~\ref{tab:svm} provides the performance of semi-supervised classification using SVMs with two different kernels --- linear and RBF. 
MAE achieves the best classification accuracy on all the settings. 
It shoulld be noted that the accuracies of SVMs with non-linear kernels fluctuate more rapidly than linear ones, particularly when the number of labeled training data is small.

\begin{table}[t]
\caption{Performance of semi-supervised classification using SVM classifiers with linear and RBF kernels. For each experiment, we report the average and standard deviation over 5 runs.}
\label{tab:svm}
\centering
{\small
\begin{tabular}[t]{l|ccc:ccc}
\hline
 & \multicolumn{3}{c:}{SVM-Linear} & \multicolumn{3}{c}{SVM-RBF} \\
\cline{2-7}
\textbf{Model} & 100 & 1000 & All & 100 & 1000 & All \\
\hline
ResNet VAE w. AF &  88.7$\pm$1.5 & 95.8$\pm$0.2 & 98.3$\pm$0.0 & 32.1$\pm$11.9 & 93.9$\pm$0.7 & 98.2$\pm$0.0 \\
VLAE & 84.9$\pm$1.8 & 94.3$\pm$0.2 & 96.7$\pm$0.0 & 51.5$\pm$5.8 & 90.9$\pm$0.4 & 97.0$\pm$0.0 \\
\hline
MAE: $\eta=1.0, \gamma=0.1$ & \textbf{91.9}$\pm$1.3 & 96.4$\pm$0.3 & 98.5$\pm$0.0 & 48.7$\pm$11.0 & 93.2$\pm$0.7 & 98.3$\pm$0.0 \\
MAE: $\eta=0.5, \gamma=0.5$ & 90.7$\pm$1.8 & 96.3$\pm$0.1 & 98.4$\pm$0.0 & 60.5$\pm$8.5 & 95.5$\pm$0.3 & 98.6$\pm$0.0 \\
MAE: $\eta=1.0, \gamma=0.5$ & 91.2$\pm$1.2 & \textbf{96.6}$\pm$0.2 & \textbf{98.5}$\pm$0.0 & \textbf{74.3}$\pm$5.8 & \textbf{96.5}$\pm$0.2 & \textbf{98.8}$\pm$0.0 \\
MAE: $\eta=2.0, \gamma=0.5$ & 90.3$\pm$1.5 & 96.5$\pm$0.1 & 98.5$\pm$0.0 & 51.5$\pm$8.9 & 92.3$\pm$1.4 & 98.5$\pm$0.0 \\
MAE: $\eta=2.0, \gamma=1.0$ & 90.3$\pm$1.5 & 96.5$\pm$0.1 & 98.5$\pm$0.0 & 54.5$\pm$9.8 & 95.0$\pm$0.5 & 98.5$\pm$0.0 \\
\hline
\end{tabular}
}
\end{table}

\subsubsection{Latent Space Visualization}\label{appendix:visual}
Figure~\ref{fig:tsne} visualize the latent spaces of VAEs and MAEs with different settings on MNIST, by t-Distributed Stochastic Neighbor Embedding (t-SNE)~\citep{maaten2008visualizing}.
The first row displays the visualizations of ResNet VAE, VLAE without free-bits training and VLAE with free-bits training.
The following three rows display visualizations of MAEs with $\gamma \in [0.1, 0.5, 1.0]$ and $\eta \in [0.5, 1.0, 2.0]$.
We see that large $\eta$ encourages the posteriors to diverse from each other, by splitting the data into small groups. 
Meanwhile, increasing $\gamma$ is effective to prevent the phenomenon.

\subsection{Experiments for CIFAR-10}\label{appendix:cifar}
Following \citet{kingma2016improved} and \citet{chenvariational}, latent codes are represented by 16 feature maps of 8x8.
Prior distribution is factorized Gaussian transformed by 8 auto-regressive flows, each of which is implemented by 3-layer masked CNNs~\citep{oord2016pixel} with 128 feature maps.
Between every other auto-regressive flow, the ordering of stochastic units is reversed.
PixelCNN++~\citep{salimans2017pixelcnn++} with 7x3 receptive field is used as the decoder.
Due to the limitation of computational resources, we used batch size 64 in our experiments. 
Same as experiments on binary images, Polyak averaging~\citep{polyak1992acceleration} was used to compute the final parameters, with $\alpha=0.999$.

\section{Generated Samples from VLAE and MAE}
\begin{figure}[h]
\centering
\begin{minipage}[t]{0.99\textwidth}
\centering
\subfloat[MNIST samples from VLAE]{
\includegraphics[width=0.45\textwidth]{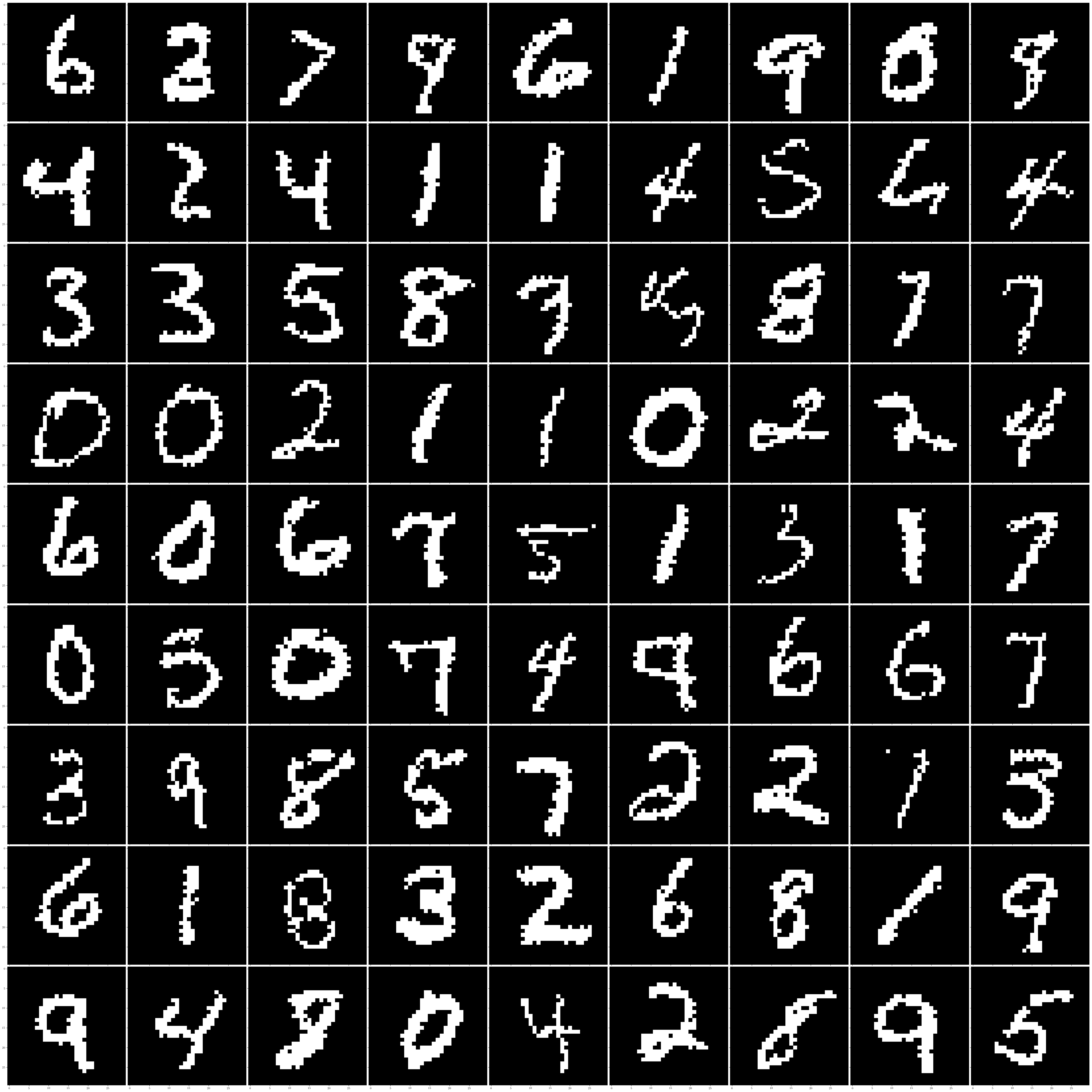}
\label{subfig:mnist_sample_vae}
}
\subfloat[MNIST samples from MAE]{
\includegraphics[width=0.45\textwidth]{figs/binary/mae_mnist_sample.png}
\label{subfig:mnist_sample_mae}
}
\\
\subfloat[OMNIGLOT samples from VLAE]{
\includegraphics[width=0.45\textwidth]{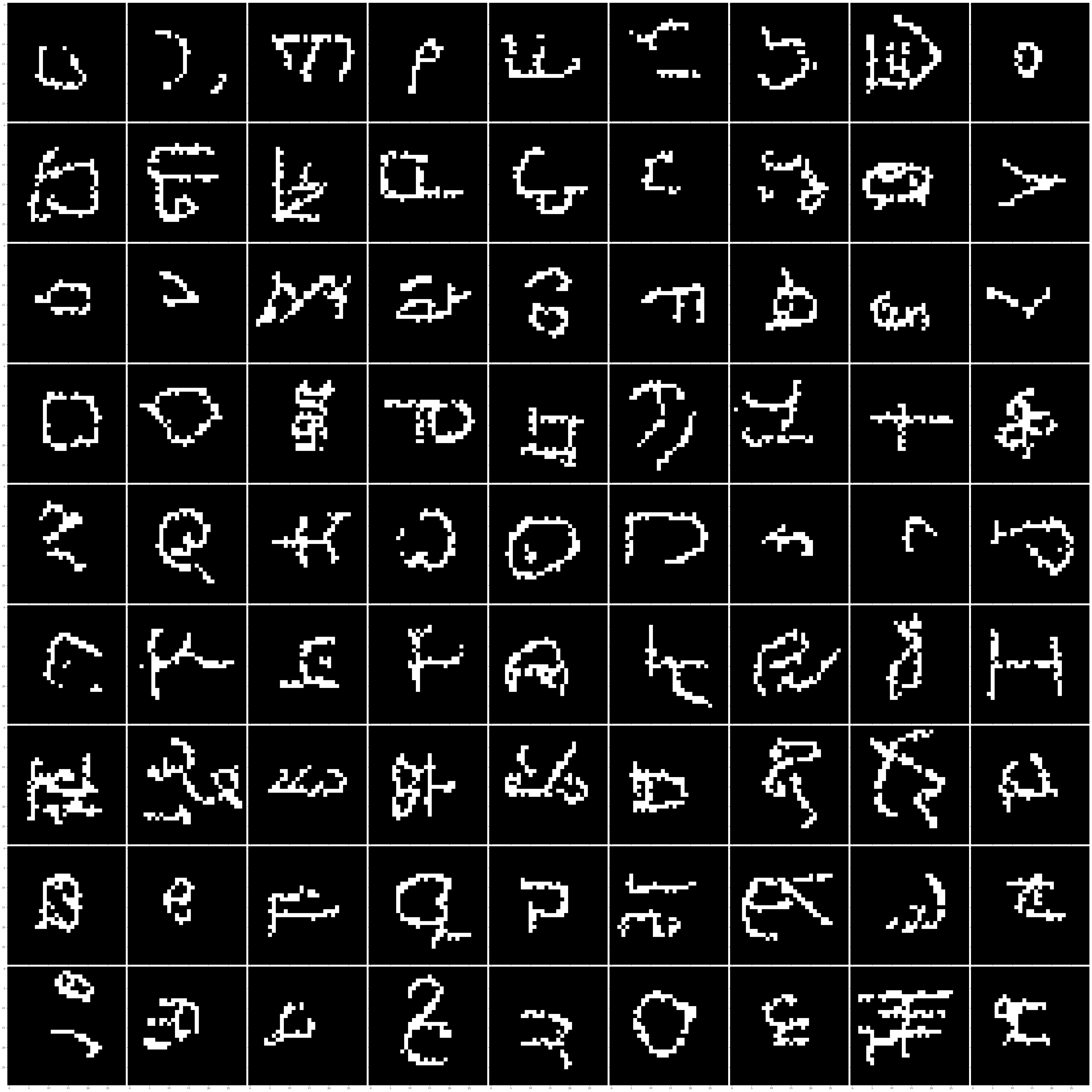}
\label{subfig:omni_sample_vae}
}
\subfloat[OMNIGLOT samples from MAE]{
\includegraphics[width=0.45\textwidth]{figs/binary/mae_omniglot_sample.png}
\label{subfig:omni_sample_mae}
}
\\
\subfloat[CIFAR-10 samples from VLAE]{
\includegraphics[width=0.47\textwidth]{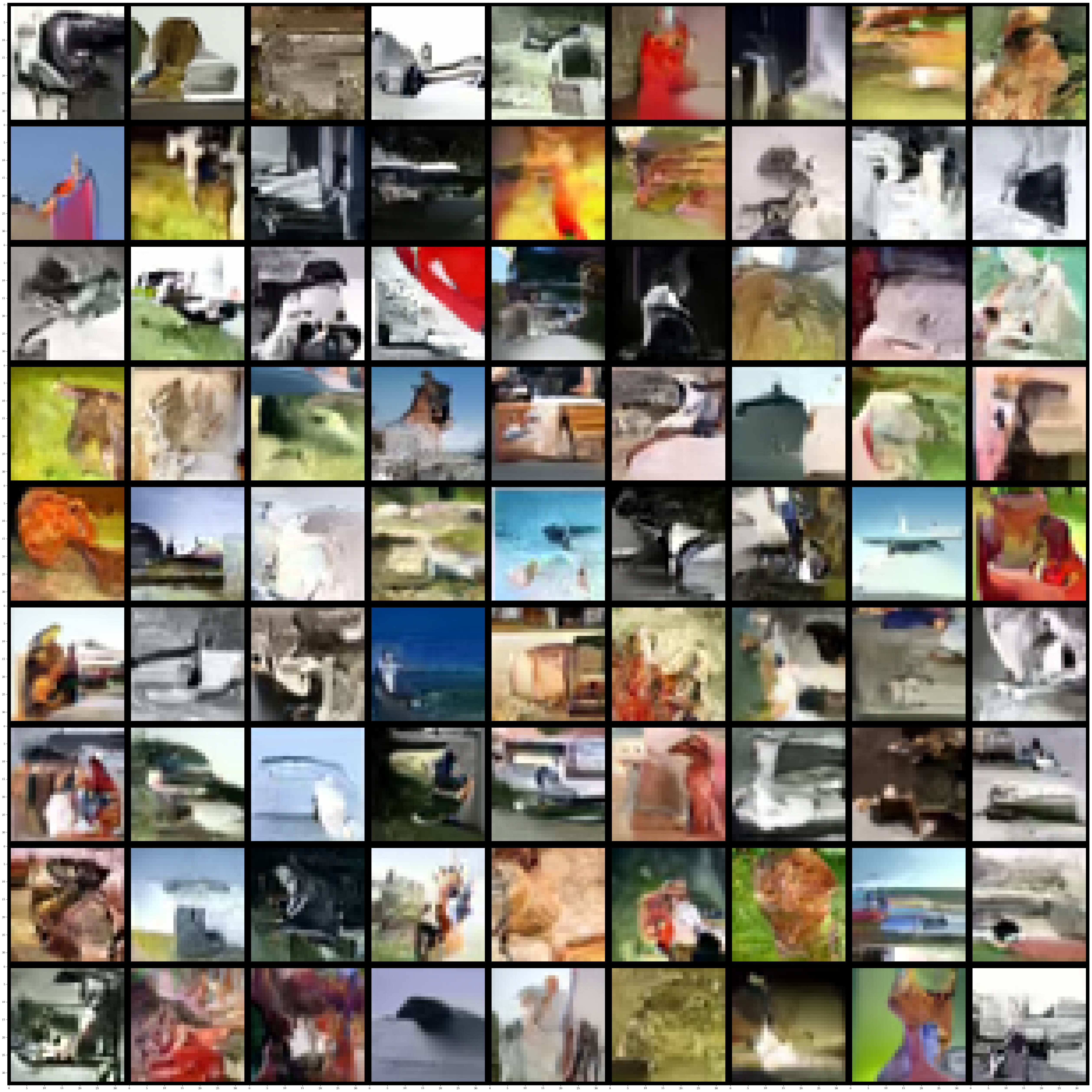}
\label{subfig:cifar_sample_vae}
}
\subfloat[CIFAR-10 samples from MAE]{
\includegraphics[width=0.47\textwidth]{figs/natural/mae_cifar10_sample.png}
\label{subfig:cifar_sample_mae}
}
\end{minipage}
\caption{Image samples from VLAE and MAE.}
\label{fig:sample}
\end{figure}

Figure~\ref{fig:sample} provides generated images on MNIST, OMNIGLOT and CIFAR-10 from VLAE and MAE.

\end{document}